\documentclass[journal]{IEEEtran}
\usepackage{times}
\usepackage{epsfig}
\usepackage{graphicx}
\usepackage{amsmath}
\usepackage{amssymb}
\usepackage{bm}
\usepackage{booktabs}
\usepackage{xcolor}

\usepackage[breaklinks=true,bookmarks=false]{hyperref}

\usepackage{etoolbox}
\makeatletter
\patchcmd{\@makecaption}
  {\scshape}
  {}
  {}
  {}
\makeatother

\newcommand{\etal}{\mbox{\emph{et al.\ }}}

\begin{document}

\title{An Unsupervised Game-Theoretic Approach to Saliency Detection}

\author{Yu~Zeng, Mengyang~Feng, Huchuan~Lu, and Ali~Borji
\thanks{Y. Zeng is with School of Information and Communication Engineering at Dalian University of Technology, Dalian, China. E-mail: zengyu@mail.dlut.edu.cn.}
\thanks{M. Feng is with School of Information and Communication Engineering at Dalian University of Technology, Dalian, China. E-mail: mengyangfeng@gmail.com}
\thanks{H. Lu is with School of Information and Communication Engineering at Dalian University of Technology, Dalian, China. E-mail: lhchuan@dlut.edu.cn. }
\thanks{A. Borji is with the Center for Research in Computer Vision and Computer Science Department at the University of Central Florida. E-mail: aliborji@gmail.com. }

\thanks{Manuscript received xx 2017.}
}
\markboth{Submitted to IEEE Transactions on Image Processing,~Vol.~xx, No.~xx, xx}%
{An Unsupervised Game-Theoretic Approach to Saliency Detection}
\maketitle
\begin{abstract}
We propose a novel unsupervised game-theoretic salient object detection algorithm that does not require labeled training data. First, saliency detection problem is formulated as a non-cooperative game, hereinafter referred to as Saliency Game, in which image regions are players who choose to be "background" or "foreground" as their pure strategies. A payoff function is constructed by exploiting multiple cues and combining complementary features. Saliency maps are generated according to each region's strategy in the Nash equilibrium of the proposed Saliency Game. Second, we explore the complementary relationship between color and deep features and propose an Iterative Random Walk algorithm to combine saliency maps produced by the Saliency Game using different features. Iterative random walk allows sharing information across feature spaces, and detecting objects that are otherwise very hard to detect. Extensive experiments over 6 challenging datasets demonstrate the superiority of our proposed unsupervised algorithm compared to several state of the art supervised algorithms. 
\end{abstract}

\section{Introduction}
Saliency detection is a preprocessing step in computer vision which aims at finding salient objects in an image~\cite{Achanta2010SLIC}. Saliency helps allocate computing resources to the most informative striking objects in an image, rather than processing the background. This is very appealing for many computer vision tasks such as object tracking, image and video compression, video summarization, image retrieval and classification. A lot of previous effort has been spent on this problem and has resulted in several methods~\cite{borji2015salient,Borji_PAMI13}. Yet, saliency detection in arbitrary images remains to be a very challenging task, in particular over images with several objects amidst high background clutter. 

\begin{figure}
\begin{center}
\label{fig:1}
 \includegraphics[width=1\linewidth]{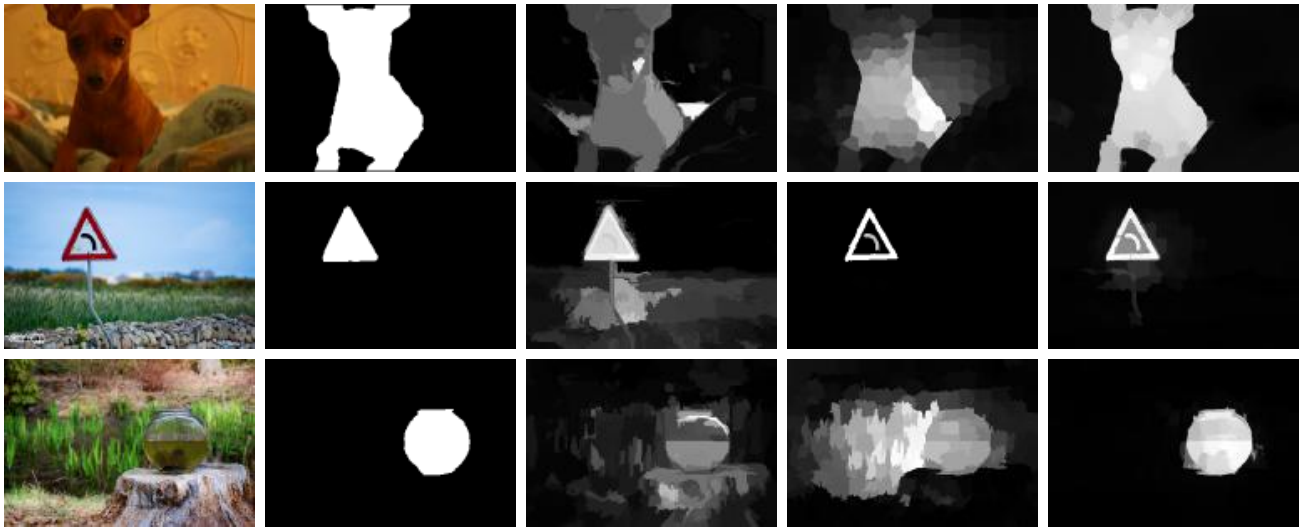}
   {~~~(a)~~~~~~~~~~~~~(b)~~~~~~~~~~~~(c)~~~~~~~~~~~~~(d)~~~~~~~~~~~~~(e)~~~}
\end{center}
\vspace{-10pt}
\caption{\small{Saliency detection results by different methods. (a) Input images, (b) Ground truth maps, (c) DRFI method~\cite{jiang2013salient}, a supervised method based on handcrafted features, (d) MR method~\cite{yang2013saliency}, an unsupervised method taking image boundary regions as background seeds, (e) Our method.}}
\end{figure}

On the one hand, unsupervised methods are usually more economical than supervised ones because no training data is needed. But they usually require a prior hypothesis about salient objects, and their performance heavily depend on reliability of the utilized prior. Take a recently popular label propagation approach as an example (e.g.,~\cite{Jiang2013Saliency}\cite{yang2013saliency}\cite{Kong2016Pattern}\cite{Wang2016GraB}). First, seeds are selected according to some prior knowledge (e.g., boundary background prior), and then, labels are propagated from seeds to unlabeled regions. They work well in most of the cases, but their results will be inaccurate if the seeds are wrongly chosen. 
For instance, when image boundary regions serve as background seed, the output will be unsatisfactory if the salient objects touch the image boundary (see the first row of Figure~\ref{fig:1}(d)).

On the other hand, supervised methods are generally more efficient. Compared with unsupervised methods based on heuristic rules, supervised methods can learn more representative properties of salient objects from numerous training images. The prime example is deep learning based methods~\cite{Wang2015Deep}\cite{zhao2015saliency}\cite{Li2015Visual}\cite{Lee_2016_CVPR}. Owning to their hierarchical architecture, deep neural networks (e.g., CNNs~\cite{Lecun1998Gradient}) can learn high-level semantically rich features. Consequently, these methods are able to detect semantically salient objects in complex backgrounds. However, off-line training a CNN needs a great deal of training data. As a result, using CNNs for saliency detection, although effective, is relatively less economical than unsupervised approaches. 

In this paper, we attempt to overcome the aforementioned drawbacks. To begin with, the saliency detection problem is formulated as a Saliency Game among image regions. Our main motivation in formulating saliency and attention in a game-theoretic manner is the very essence of attention which is the competition among objects to enter high level processing. Most previous methods formulate saliency detection as minimizing one single energy function that incorporates saliency priors. Different image regions are often considered through adding terms in the energy function (e.g.,~\cite{zhu2014saliency}) or sequentially (e.g.,~\cite{yang2013saliency}). If the priors are wrong, optimization of their energy function might lead to wrong results. In contrast, we define one specific payoff function for each superpixel which incorporates multiple cues including spatial position prior, objectness prior, and support from others. Adopting two independent priors makes the proposed method more robust since when one prior is inappropriate, the other might work. The goal of the proposed Saliency Game is to maximize the payoff of each player given other players’ strategies. This can be regarded as maximizing many competing objective functions simultaneously. The game equilibrium automatically provides a trade-off, so that when some image region can not be assigned a right saliency value by optimizing one objective function (e.g., due to misleading prior), optimization of the other objective functions might help to give them a right saliency value. This approach seems very natural for attention modeling and saliency detection, as also features and objects compete to capture our attention.  

\begin{figure*}
\setlength{\abovedisplayskip}{0pt}
\setlength{\belowdisplayskip}{0pt}
\begin{center}
 \includegraphics[width=1\linewidth]{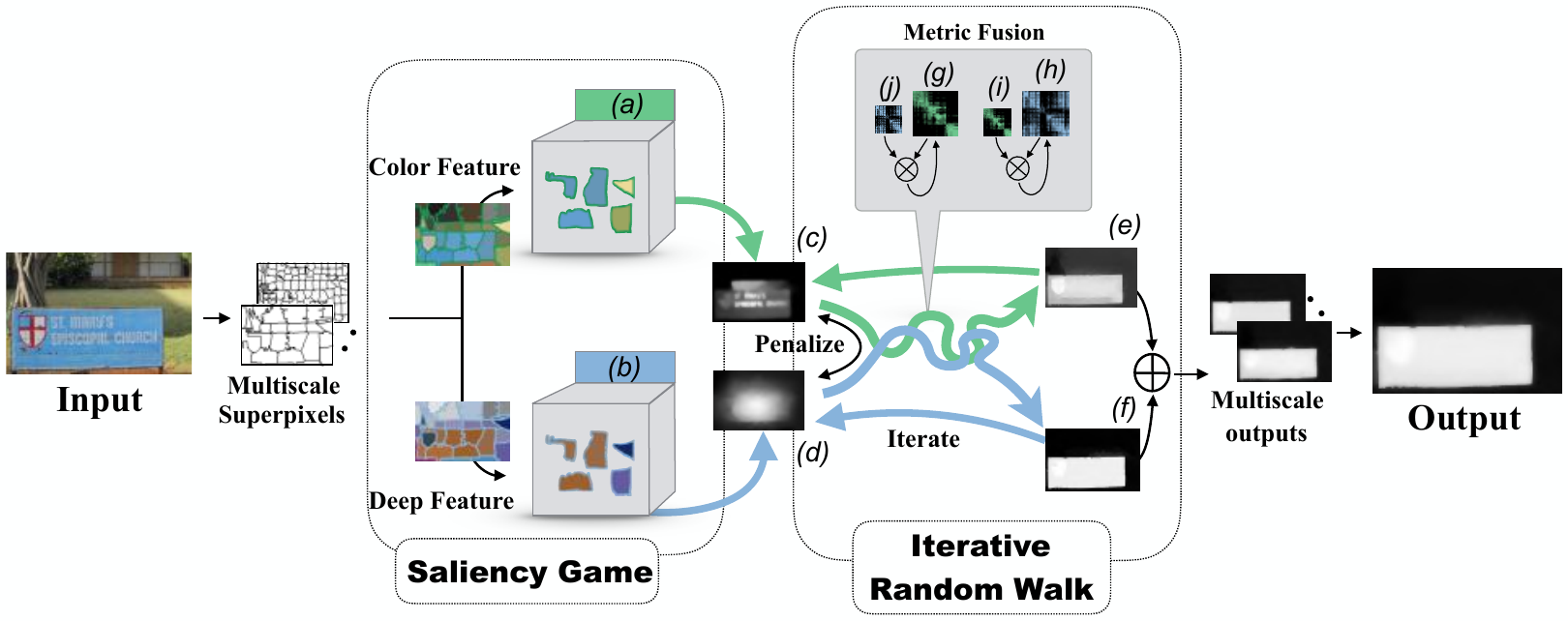}
\end{center}
\vspace{-10pt}
   \caption{\small{The pipeline of our algorithm. The input image is segmented into superpixels in several scales. Th following process is applied to each scale, and the average of the results in all scales is taken as the final saliency map. First, the \textit{Saliency Game} among superpixels is formulated in two feature spaces ($(a)$ and $(b)$), respectively to generate corresponding results ($(c)$ and $(d)$).
   Second, an \textit{Iterative Random Walk} is constructed to refine the results of the Saliency Game. Then outputs $(e)$ and $(f)$ are summed (with weights) as the result in one scale. In the Iterative Random Walk, complete affinity metric ($(g)$ in deep feature space and $(h)$ in color space) in one feature space is fused with neighboring affinity metric ($(i)$ in color space and $(j)$ in deep space) in another feature space. This is done to exploit the complementary relationship between the two feature spaces. Finally, we average the results in all scales to form the final saliency map. 
}}
\label{fig:2}
\end{figure*}

In addition, it is known that one main factor for the astonishing success of deep neural networks is their powerful ability to learn high-level semantically-rich features. Using features extracted from a pre-trained CNN to build an unsupervised method seems a considerable option, as it allows utilizing the aforementioned strength while avoiding time-consuming training. However, rich semantic information comes with the cost of diluting image features through convolution and pooling layers. Due to this, we also use traditional color features as supplementary information. To make full use of these two complementary features for better detection results, we avoid taking the weighted sum of the raw results generated by the above Saliency Game in the two feature spaces. Instead, we further propose an Iterative Random Walk algorithm across two feature spaces, deep features and the traditional CIE-Lab color features, to refine saliency maps. In every iteration of the Iterative Random Walk, the propagation in the two feature spaces are penalized by the latest output of each other. Figure~\ref{fig:2} shows the pipeline of our algorithm. 

In a nutshell, the main contributions of our work include: 
\begin{enumerate}
\item We propose a novel unsupervised Saliency Game to detect salient objects. Adopting two independent priors improves robustness. The nature of game equilibria assures accuracy when both priors are unsatisfactory,
\item Utilizing semantically-rich features extracted from pre-trained fully convolutional networks (FCNs)~\cite{Long2015Fully}, in an unsupervised manner, the proposed method is able to identify salient objects in complex scenes, where traditional methods that use handcrafted features may fail (see Figure~\ref{fig:1}(c)), and
\item An Iterative Random Walk algorithm across two feature spaces is proposed that takes advantage of the complementary relationship between the color feature space and the deep feature space to further refine the results. 
\end{enumerate}

\section{Related work}

Some saliency works have followed an unsupervised approach. In~\cite{Jiang2013Saliency}, saliency of each region was defined as its absorbed time from boundary nodes, which measures its global similarity with all boundary regions. 
Yang \etal.~\cite{yang2013saliency} ranked the similarity of image regions with foreground or background cues via graph-based manifold ranking. Saliency value of each image element was determined based on its relevance to given seeds. 
In~\cite{Kong2016Pattern}, saliency pattern was mined to find foreground seeds according to prior maps. Foreground labels were propagated to unlabeled regions. Tong \etal.~\cite{tong2015bootstrap} proposed a learning algorithm to bootstrap training samples generated from prior maps. These methods exploited either boundary background prior or foreground prior from a prior map, while we adopt two different priors in our method for robustness purposes. Priors only act as weak guidance with very small weights in the payoff function of our proposed Saliency Game. 

Some deep learning based saliency detection methods have achieved great performance. In~\cite{Wang2015Deep}, two deep neural networks were trained, one to extract local features and the other to conduct a global search. Zhao \etal.~\cite{zhao2015saliency} proposed a multi-context deep neural network taking both global and local context into consideration. Li \etal.~\cite{Li2015Visual} explored high-quality visual features extracted from deep neural networks to improve the accuracy of saliency detection. In \cite{Lee_2016_CVPR}, high level deep features and low level handcrafted features were integrated in a unified deep learning framework for saliency detection. All above methods needed a lot of time and many images for training. In this work, we are not against the CNN models, but we combine deep features with traditional color features in an unsupervised way, which result in an efficient unsupervised method complementary to CNNs that does on par with the above models that need labeled training data. Hopefully, this will encourage new models that can utilize both labeled and unlabeled data. 

Furthermore, there are many computer vision and learning tasks in which game theory has been applied successfully. A grouping game among data points was proposed in~\cite{Torsello2006Grouping}. Albarelli \etal.~\cite{Albarelli2009Matching} proposed a non-cooperative game between two sets of objects to be matched. A game between a region-based segmentation model and a boundary-based segmentation model was proposed in~\cite{Chakraborty1999Game} to integrate two sub-modules. Erdem \etal~\cite{Erdem2012Graph} formulated a multi-player game for transduction learning, whereby equilibria correspond to consistent labeling of the data. In~\cite{Miller1991Copositive}, Miller \etal showed that the relaxation labeling problem~\cite{Hummel1983On} is equivalent to finding Nash equilibria for polymatrix n-person games. However, to the best of our knowledge, game theory has not yet been used for salient object detection. 

\section{Definitions and symbols}
\label{def}
In this section, we introduce definitions and symbols that will be  used throughout the paper. 

\noindent \textbf{Superpixels:}
In our model, the processing units are superpixels segmented from an input image by the SLIC algorithm~\cite{Achanta2010SLIC}. $\mathcal{I} = \{1, 2, 3, ..., N\}$ denotes the enumeration of the set of superpixels.  $P_i(x,y)$, $i \in \mathcal{I}$, denotes the mask of the $i$-th superpixel, where $P_i(x, y) = 1$ indicates that the pixel located at $(x, y)$ of the input image belongs to the $i$-th superpixel, and $P_i(x, y) = 0$, otherwise.

\noindent \textbf{Features:}
We use FCN-32s~\cite{Long2015Fully} features due to its great success in semantic segmentation. We choose the output of the \texttt{Conv5} layer as feature maps for the input image. This is because features in the last layers of CNNs encode semantic abstraction of objects and are robust to appearance variations. Since the feature maps and the image are not of the same resolution, we resize the feature maps to the input image size via linear interpolation. We denote the affinity between the $i$-th superpixel and the $j$-th superpixel in deep feature space as $A^d(i,j)$, which is defined to be their Gaussian weighted Euclidean distance:
\begin{equation}
\setlength{\abovedisplayskip}{2pt}
\setlength{\belowdisplayskip}{2pt}
A^d(i, j) = \exp \big( -\frac{\left\| \boldsymbol{f}_i^d - \boldsymbol{f}_j^d \right\|^2}{\sigma^2} \big), 
\end{equation}
where $\boldsymbol{f}^d_i$ is the deep feature vector of superpixel $i$. Each superpixel is represented by the mean deep feature vector of all its contained pixels.

The semantically-rich deep features can help accurately locate the targets but fail to describe the low-level information. Therefore, we also employ color features as a complement to deep features. Inspired by~\cite{Jiang2013Saliency}, we use CIE-Lab color histograms to describe superpixels' color appearance. With CIE-Lab color space divided into $8^3$ ranges, the color feature vector of the $i$-th superpixel is denoted as $\boldsymbol{f}_i^c$.
Affinity between superpixels $i$ and $j$ in the color feature space is denoted as $A^c(i,j)$, which is defined to be their Gauss weighted Chi-square distance:
\begin{equation}
\setlength{\abovedisplayskip}{2pt}
\setlength{\belowdisplayskip}{2pt}
A^c(i, j) = \exp \big( -\frac{\chi^2(\boldsymbol{f}_i^c,\boldsymbol{f}_j^c )}{\sigma^2} \big). 
\end{equation}
\noindent\textbf{Neighbor:}
We adopt a definition of 2-hoop neighbor which is frequently used in superpixel based saliency detection methods. The set of the $i$-th superpixel's neighbors is denoted as $ \mathcal{N}(i) = \mathcal{N}_1(i) \cup \mathcal{N}_2(i) \cup \mathcal{N}_3(i)$, where $\mathcal{N}_1(i)$ indicates the set of superpixels who share at least one common edge with the $i$-th superpixel. $\mathcal{N}_2(i)$ and $\mathcal{N}_3(i)$ are defined as follows: 
\begin{equation}
\setlength{\abovedisplayskip}{2pt}
\setlength{\belowdisplayskip}{2pt}
\mathcal{N}_2(i) = \{j | j \in \mathcal{N}_1(k), k \in \mathcal{N}_1(i), j \ne i\},
\end{equation}
\begin{equation}
\mathcal{N}_3(i) =
\begin{cases}
\emptyset &\mbox{if}~i \notin \mathcal{B}\\
\{j | j \in \mathcal{B}, j \ne i\} &\mbox{if}~i \in \mathcal{B}
\end{cases},
\end{equation}
where $\mathcal{B}$ denotes the set of superpixels in image boundary. 

\section{Saliency game}\label{section:my}
Here, we formulate a non-cooperative game among superpixels to detect salient objects in an input image. The input image is firstly segmented into $N$ superpixels which act as players in the Saliency Game. Each player chooses to be "background" or "foreground" as its pure strategy and its mixed strategy corresponds to this superpixel's saliency value. After showing their strategies, players obtain some payoff according to both their own and other players' strategies. Payoff is determined by a payoff function which incorporates position and objectness cues as well as support from others. We use each player's mixed strategy in the Nash equilibrium of the proposed Saliency Game as the saliency value of this superpixel in the output saliency map. Such an equilibrium corresponds to a steady state where each player plays a strategy that maximize its own payoff when the remaining players' strategies are kept fixed, which provides a globally plausible saliency detection result.

\subsection{Game setting}\label{gamesetting}
The pure strategy set is denoted as $\mathcal{S} = \{0, 1\}$, indicating "to be foreground" or "to be background", respectively. All superpixels' pure strategies are collectively called a pure strategy profile, denoted as $\boldsymbol{s} = (s_1, ..., s_N)$. The strategy profile set is denoted as $\Theta$. $\pi_{ij}(s_i, s_j)$ denotes a single payoff that superpixel $i$ obtains, when playing pure strategy $s_i$ against superpixel $j$ who holds a pure strategy $s_j$, in their 2-person game. There are four possible values for $\pi_{ij}(s_i, s_j)$ that can be put into a $2\times2$ matrix
$B_{ij}$, 
\begin{equation}
\setlength{\abovedisplayskip}{2pt}
\setlength{\belowdisplayskip}{2pt}
\label{pmatrix}
B_{ij} = \begin{pmatrix} \pi_{ij}(0, 0) & \pi_{ij}(0, 1) \\ \pi_{ij}(1, 0) &  \pi_{ij}(1, 1) \end{pmatrix} \\.
\end{equation}
Payoff of superpixel $i$ in pure strategy profile $\boldsymbol{s}$, where $\forall j \in \mathcal{I}$ the $j$-th superpixel's pure strategy $s_j$ is the $j$-th component of vector $\boldsymbol{s}$, is denoted as $\pi_i(\boldsymbol{s})$. Payoff of superpixel $i$ when it adopts a pure strategy $t_i$ (not necessarily the $i$-th component of $\boldsymbol{s}$), while all other superpixels adopt pure strategies in pure strategy profile $\boldsymbol{s}$ is denoted as $\pi_{i}(t_i, \boldsymbol{s}_{-i})$. We make an assumption that the total payoff of superpixel $i$ for playing with all others is the summation of payoffs for playing 2-player games with every other single superpixel. Formally, we assume that $\pi_{i}(\boldsymbol{s}) = \sum_{j \neq i} \pi_{ij}(s_i, s_j)$ and $\pi_{i}(t_i, \boldsymbol{s}_{-i}) = \sum_{j \neq i} \pi_{ij}(t_i, s_j)$.  

A pure best reply for player $i$ against a pure strategy profile $\boldsymbol{s}$ is a pure strategy such that no other pure strategy gives a higher payoff to $i$ against $\boldsymbol{s}$. The i-th player's pure best-reply correspondence, which maps each pure strategy profile $ \boldsymbol{s} \in \Theta$ to a pure strategy $s_i \in \mathcal{S}$, is denoted as $\beta_i: \Theta \rightarrow \mathcal{S}$: 
\begin{equation}
\setlength{\abovedisplayskip}{2pt}
\setlength{\belowdisplayskip}{2pt}
\beta_i(\boldsymbol{s}) = \{ s_i \in \mathcal{S} | \pi_i(s_i, \boldsymbol{s}_{-i}) \ge \pi_i(t_i, \boldsymbol{s}_{-i}), \forall t_i \in \mathcal{S} \}. 
\end{equation}
The combined pure best-reply correspondence $\beta: \Theta \rightarrow \Theta$ is defined as the cartesian product of all players' pure best-reply correspondence:
\begin{equation}
\setlength{\abovedisplayskip}{2pt}
\setlength{\belowdisplayskip}{2pt}
\label{equilibria}
\beta(\boldsymbol{s}) = \times_{i\in\mathcal{I}} \beta_i(\boldsymbol{s}) \subset \Theta. 
\end{equation}
A pure strategy profile $\boldsymbol{s}$ is a pure Nash equilibrium if $\boldsymbol{s} \in \beta(\boldsymbol{s})$.

A probability distribution over the pure strategy set is termed as a mixed strategy. Mixed strategy of the $i$-th superpixel is denoted as a 2-dimensional vector $\boldsymbol{z}_i = (z_{i}^0, z_{i}^1)^{\rm T}$, while $z_{i}^0 = P(s_i=0), z_{i}^1 = P(s_i = 1)$ and $z_{i}^0 + z_{i}^1 = 1$. The set of mixed strategies is denoted as $\Delta$. A pure strategy thereby can be regarded as an extreme mixed strategy where only one component is 1 and the other one is 0, e.g., $i$-th player's pure strategy $s_i = 1$ is equivalent to its mixed strategy $\boldsymbol{z}_i = (0, 1)^{\rm T}$ because $P(s_i=0)=0$ and $P(s_i=1)=1$. Correspondingly, expected payoff of superpixel $i$ for playing mixed strategy $\boldsymbol{z}_i$ against superpixel $j$ holding mixed strategy $\boldsymbol{z}_j$ is denoted as $u_{ij}(\boldsymbol{z}_i, \boldsymbol{z}_j) = \boldsymbol{z}_i^{\rm T} B_{ij} \boldsymbol{z}_j$. We also denote $Z = (\boldsymbol{z}_1, ..., \boldsymbol{z}_N)$, $u_i(Z) = \sum_{j \neq i} u_{ij}(\boldsymbol{z}_i, \boldsymbol{z}_j)$ and $u_i(\boldsymbol{w}_i, Z_{-i}) = \sum_{j \neq i} u_{ij}(\boldsymbol{w}_i, \boldsymbol{z}_j)$ to be mixed strategy version of $\boldsymbol{s}$, $\pi_{i}(\boldsymbol{s})$ and $\pi_{i}(t_i, \boldsymbol{s}_{-i})$. Similarly, a mixed Nash equilibrium is also defined to be a mixed strategy profile which is a mixed best reply to itself. These symbols or definitions are not stated here individually due to limited space. 

From the definition of the Nash equilibrium above, it can be inferred that in a Nash equilibrium of a game, each player adopts a strategy that maximizes its own payoff when other players' strategies are fixed.

\subsection{Payoff function}
We have assumed in Section~\ref{gamesetting} that the total payoff of superpixel $i$ for playing with all others is the summation of every single payoff in its 2-person games with every other superpixel. Hence, here we focus on modeling payoff of every 2-person game. We define the payoff $\pi_{ij}(s_i, s_j)$ of superpixel $i$ for its 2-person game with $j$ as a weighted sum of three terms:
\begin{equation}
\setlength{\abovedisplayskip}{2pt}
\setlength{\belowdisplayskip}{2pt}
\label{payoff}
\pi_{ij}(s_i, s_j) = \lambda_1 \cdot \mbox{pos}_i(s_i) + \lambda_2 \cdot \mbox{obj}_i(s_i) + \mbox{spt}_{ij}(s_i, s_j), 
\end{equation}
where $\mbox{pos}_i(s_i)$, $\mbox{obj}_i(s_i)$, and $\mbox{spt}_{ij}(s_i, s_j)$ indicate the $i$-th superpixel's position prior, objectness prior and support that superpixel $j$ gives to superpixel $i$, respectively.
$\lambda_1$ and $\lambda_2$ 
are parameters controlling the weight of the first two terms. 

\noindent \textbf{Position:} Position prior term in the payoff function is formulated based on the observation that salient objects often fall at the image center. The position term should give a greater payoff when, a) Center superpixels choose to be foreground,
and b) Boundary superpixels choose to be background. Assuming $(x_0, y_0)$ to be the image center, and $(x_i, y_i)$ to be the center coordinate of superpixel $i$, the position prior term is defined as follows,
\begin{equation}
\setlength{\abovedisplayskip}{2pt}
\setlength{\belowdisplayskip}{2pt}
\label{pos}
\mbox{pos}_i(s_i) =
\begin{cases}
\frac{1}{N} \exp\{-\frac{(x_i - x_0)^2 + (y_i - y_0)^2}{\sigma}\} & \mbox{if}~s_i = 1 \\
\frac{1}{N} (1 - \exp\{-\frac{(x_i - x_0)^2 + (y_i - y_0)^2}{\sigma}\}) &\mbox{if}~s_i = 0
\end{cases}. 
\end{equation}

\noindent \textbf{Objectness:} Generally, objects attract more attention than background clutter. Hence superpixels which are part of an object are more likely to be salient. The objectness term should give a greater payoff when, a) Superpixels with high objectness choose to be foreground, and b) Superpixels with low objectness choose to be background.

We exploit the geodestic object proposal (GOP) ~\cite{Kr2014Geodesic} method to extract a set of object segmentations, and define the objectness of a superpixel according to its overlap with all GOP proposals as follows: 
\begin{equation}
\setlength{\abovedisplayskip}{2pt}
\setlength{\belowdisplayskip}{2pt}
\small
\label{obj}
\mbox{obj}_i(s_i)=
\begin{cases}
\frac{1}{N \cdot N_o} \sum_{j=1}^{N_o} \frac{\sum_{x, y} O_j(x,y) \times P_i(x,y)}{\sum_{x,y} P_i(x,y)}, ~~~~\mbox{if}~s_i = 1 \\ \\
\frac{1}{N} (1 - \frac{1}{N_o} \sum_{j=1}^{N_o} \frac{\sum_{x, y} O_j(x,y) \times P_i(x,y)}{\sum_{x,y} P_i(x,y)}),~\mbox{otherwise}
\end{cases}
\end{equation}
where $\{O_j\}_{N_o}$ is the set of object candidate masks generated by the GOP method, where $O_j(x, y) = 1$ indicates that the pixel located at $(x, y)$ of the input image belongs to the $j$-th object proposal, and $O_j(x, y) = 0$, otherwise. ${P_i}$ is the mask of the $i$-th superpixel as in Section~\ref{def}. 

\noindent \textbf{Support:} With a much larger weight in the payoff function ($\lambda_1$ and $\lambda_2$ being small), support from others is the main source of payoff obtained by each superpixel. When playing with an opponent, each superpixel judges if the opponent's strategy is right or wrong with its own stance, and provides a higher or lower even negative support to the opponent accordingly. More precisely,
\begin{itemize}
   \item Each superpixel takes a neutral attitude to opponents who hold different pure strategies from itself, and provides them zero support.
   \item If an opponent adopts the same pure strategy as superpixel $i$, 
   \begin{itemize}
      \item if the opponent's strategy is similar to it, then superpixel $i$ provides the opponent a great support in recognition of its choice. 
      \item else if the opponent is not similar with it, then superpixel $i$ provides the opponent a small even negative support as punishment.
   \end{itemize}
\end{itemize}

Formally, the support term is defined as follows,
\begin{equation}
\setlength{\abovedisplayskip}{2pt}
\setlength{\belowdisplayskip}{2pt}
\label{supp}
\mbox{spt}_{ij}(s_i, s_j) =
\begin{cases}
A(i, j) -  \frac{\alpha}{N} \sum_{k=1}^N A(i, k), & \mbox{if}~s_i = s_j \\
0, & \mbox{if}~s_i \ne s_j
\end{cases}, 
\end{equation}
where $\alpha$ is a positive constant, $A(i, j)$ is the affinity between superpixels $i$ and $j$, defined as $A^c$ and $A^d$ in Section~\ref{def}. 

So far, we have modeled the payoff $\pi_i(s_i, s_j)$ that superpixel $i$ obtains by playing a 2-person pure strategy game with superpixel $j$. The expected payoff $u_i(\boldsymbol{w}_i,Z_{-i})$ that superpixel $i$ obtains by playing mixed strategy game with all others adopting strategies in mixed strategy profile $Z$ can be given based on the definition stated at the beginning of this section, 
\begin{equation}
\setlength{\abovedisplayskip}{0.1pt}
\setlength{\belowdisplayskip}{0.1pt}
u_i(\boldsymbol{w}_i,Z_{-i}) = \sum_{j=1 \land j \ne i}^N \boldsymbol{w}_i^{\rm T} B_{ij} \boldsymbol{z}_j. 
\end{equation}

\begin{figure}[t]
\begin{center}
   \includegraphics[width=1\linewidth]{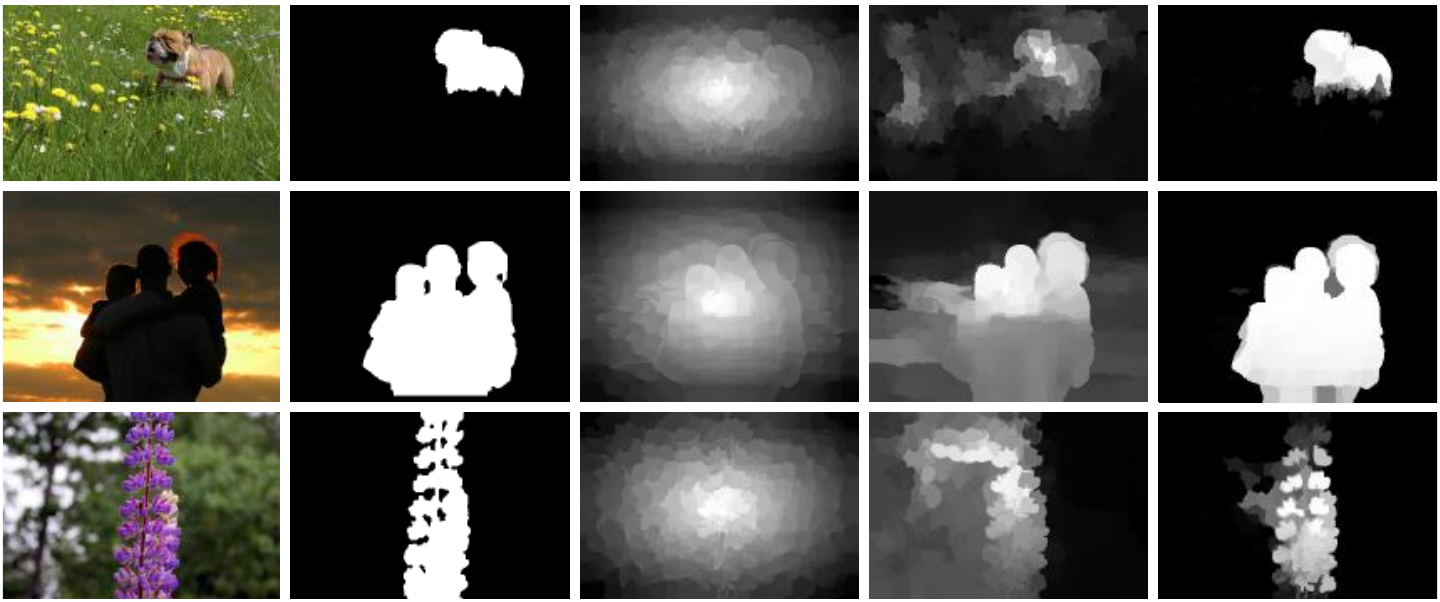}
         {~~~(a)~~~~~~~~~~~~(b)~~~~~~~~~~~~~(c)~~~~~~~~~~~~~(d)~~~~~~~~~~~~(e)~~~}
\end{center}
\vspace{-10pt}
   \caption{The proposed Saliency Game still works well even when position prior and objectness prior are not very satisfying. (a) Input images. (b) Ground Truth maps. (c) and (d) Illustration of the position term and objectness term in Eqn.~\ref{pos}. (e) Saliency maps by Saliency Game using the color feature.
   }
\label{fig:4}
\end{figure}

\subsection{Computing equilibrium}
We use Replicator Dynamics~\cite{Taylor1978Evolutionarily} to compute the mixed strategy Nash equilibrium of the proposed Saliency Game. In Replicator Dynamics, a population of individuals play the game, generation after generation. A selection process acts on the population, causing the number of users holding fitter strategies to grow faster. We use discrete time Replicator Dynamics to find the equilibrium of the game, iterating until $\forall i \in \mathcal{I}, \left |\boldsymbol{z}_i(t) - \boldsymbol{z}_i(t-1)\right| < \epsilon$,
\begin{equation}
\setlength{\abovedisplayskip}{2pt}
\setlength{\belowdisplayskip}{2pt}
\boldsymbol{z}_{i}^h(t+1) = \boldsymbol{z}_{i}^h(t) \frac{const + u_i(\boldsymbol{e}^h, {Z(t)}_{-i})}{const + u_i(Z(t))}, 
\end{equation}
where $z_{i}^h(t)$ represents the $h$-th component of the $i$-th player's mixed strategy at time $t$, $\boldsymbol{e}^h$ is a vector whose $h$-th component is 1, while other components are 0. We set the initial mixed strategies of player $i$ to $\boldsymbol{z}_i(0) = (0.5, 0.5), \forall i \in \mathcal{I}$. $const$ is background birthrate for an individual, which is set to a positive number to make sure $const + u_i(\boldsymbol{e}^h, {Z(t)}_{-i})$ is positive for all $i$ in $\mathcal{I}$~\cite{Weibull1999Eolutionary}. There could be multiple equilibria in a game, likewise in the proposed Saliency Game. Replicator Dynamics might reach different Nash equilibria if the initial state $\boldsymbol{z}_i(0)$ is set to different interior points of $\Delta$. Empirically, we find that $\boldsymbol{z}_i(0)=(0.5, 0.5)$ is a good initialization leading to plausible saliency detection. 

In the proposed Saliency Game, each superpixel inspects strategies of all other superpixels and takes a stance by providing large or small even negative support. Usually, no matter what strategy a superpixel adopts, there are both protesters and supporters. Game equilibira provide a good trade off among different influences. Thus, in the equilibrium of the proposed Saliency Game with payoff function as defined in Eqn.~\ref{payoff}, each superpixel chooses a strategy that suits itself best given its position, objectness, and support from others. Doing so has two advantages: 1) the center position prior and the objectness prior are almost independent, when one prior is unsatisfactory, the other may work.

As shown in the first row of Figure~\ref{fig:4}, the little pug appears away from image center, but since it is the only object in the image, the objectness prior identifies it correctly. 2) the two priors only serve as weak guidance and obtain small weights in the payoff function. Even when they are both unsatisfactory on some image regions, pressure from peers will impel these regions to get proper saliency values in the equilibrium of the game. As shown in the second row of Figure~\ref{fig:4}, although only heads of the people are high in position prior and objectness prior, the produced saliency map can highlight the entire object. From the third row of Figure~\ref{fig:4}, we can see that the proposed algorithm also suppresses background effectively when prior highlights background areas by mistake. Note that in order to illustrate effectiveness of the proposed Saliency Game, only color feature is used in the three shown cases.

\section{Iterative random walk}
\begin{figure}[t]
\begin{center}
  \includegraphics[width=1\linewidth]{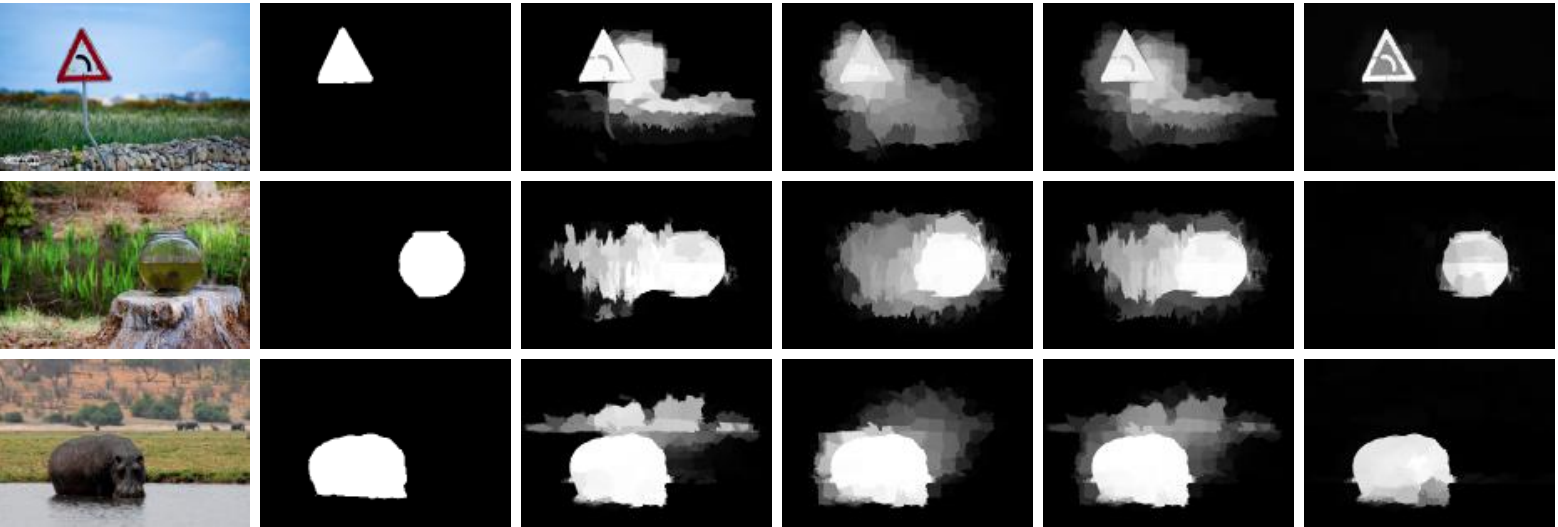}
         {~~~(a)~~~~~~~~~~(b)~~~~~~~~~~(c)~~~~~~~~~~(d)~~~~~~~~~~(e)~~~~~~~~~~(f)~~~}
\end{center}
\vspace{-10pt}
   \caption{Effect of the Iterative Random Walk. (a) Input images. (b) Ground Truth maps. (c) Saliency maps generated by our Saliency Game in the color feature space. (d) Saliency maps generated by our Saliency Game in the deep feature space. (e) The weighted summation of (c) and (d). (f) Saliency maps after refinement by the Iterative Random Walk proposed in this section. }
\label{fig:5}
\end{figure}

Traditional color features are of high-resolution, so saliency maps generated in color space are detailed and with clear edges. But due to lack of high-level information, sometimes they fail to locate the targets accurately (see Figure~\ref{fig:5}(c)). On the contrary, since deep features encode high-level concept of objects well, saliency maps generated in the deep feature space are able to find correct salient objects in an image. But due to several layers of convolution and pooling, these features are too coarse. Thus the generated saliency maps are indistinctive, as shown in Figure~\ref{fig:5}(d).

Accordingly, here, we use both complementary features for a better result. However, as shown in Figure~\ref{fig:5}(e), although the weighted sum of the two is slightly better, they are not satisfactory. To solve this problem, in this section, inspired by the metric fusion presented in~\cite{Tu2012Unsupervised}, we propose an Iterative Random Walk method to best exploit this two complementary feature spaces. In the proposed Iterative Random Walk, metrics in the two feature spaces are fused as stated in~\cite{Tu2012Unsupervised} (corss fusion in Eqn.~\ref{tu} is the work of Tu~\etal.), in addition, we also make the two propagation penalized by the latest output of each other (cross propagation in Eqn.~\ref{our1} and Eqn.~\ref{our2} is  our work). Figure~\ref{fig:6} shows that both cross fusion and cross penalization contribute.

With superpixels as nodes, a neighbor graph and a complete graph are constructed in both feature space (deep and color features). The affinity between two superpixels is assigned to the edge weight. Four weight matrices are defined:
\begin{itemize}
	\item $\mathcal{W}^d$ and $W^d$: weight matrices of neighbor and complete graphs in the deep feature space, respectively.
	\item $\mathcal{W}^c$ and $W^c$: weight matrices of neighbor and complete graphs in the color space, respectively.
\end{itemize}
In the complete graphs, there is an edge between every pair of nodes, while in the neighbor graphs, each node is connected only to its neighbors. $\forall i \in \mathcal{I}$, $\mathcal{W}^d$ and $W^d$ are defined as follows,
\begin{equation}
\setlength{\abovedisplayskip}{2pt}
\setlength{\belowdisplayskip}{2pt}
W^d(i, j) = A^d(i, j), \forall j \in \mathcal{I}, 
\end{equation}
\begin{equation}
\mathcal{W}^d(i, j) = \begin{cases}
A^d(i, j) &\mbox{if}~j \in \mathcal{N}(i) \\
0 &\mbox{if}~j \notin \mathcal{N}(i)
\end{cases}, 
\end{equation}
$\mathcal{W}^c$ and $W^c$ are defined similarly but using $A^c$. See Section~\ref{def} for definitions of $A^d$ and $A^c$. 

Firstly, let $P_{(0)}(i, j) = W(i, j)/\sum_{j=1}^N W(i, j)$, $\mathcal{P}_{(0)}(i, j) = \mathcal{W}(i, j)/\sum_{j=1}^N \mathcal{W}(i, j)$, and $t$ to be the number of iterations. Symbols with superscript $c$ or $d$ correspond to variables in the color or deep feature space, respectively. Following~\cite{Tu2012Unsupervised}, we fuse these four affinity matrices as follows,
\begin{equation}
\setlength{\abovedisplayskip}{2pt}
\setlength{\belowdisplayskip}{2pt}
\label{tu}
\begin{cases}
P^d_{(t+1)} = \mathcal{P}^c \times P^d_{(t)} \times \mathcal{P}^c\\
P^c_{(t+1)} = \mathcal{P}^d \times P^c_{(t)} \times \mathcal{P}^d
\end{cases}. 
\end{equation}

Then, using the fused affinity matrices, we let the propagation results in the two feature spaces penalize each other. Two random walk energy functions are defined as follows,
\begin{equation}
\setlength{\abovedisplayskip}{2pt}
\setlength{\belowdisplayskip}{2pt}
\small
\label{our1}
E^d_{(t+1)}(\boldsymbol{l}) = \sum_{i,j} P^d_{(t)}(i,j) (l_i - l_j)^2 + \beta \sum_{i=1}^N (l_i - l^c_{i(t)})^2, 
\end{equation}
\begin{equation}
\setlength{\abovedisplayskip}{2pt}
\setlength{\belowdisplayskip}{2pt}
\small
\label{our2}
E^c_{(t+1)}(\boldsymbol{l}) = \sum_{i,j} P^c_{(t)}(i,j) (l_i - l_j)^2 + \beta \sum_{i=1}^N (l_i - l^d_{i(t)})^2. 
\end{equation}
\noindent where $\boldsymbol{l}$ is the label vector, $l_i$ is the $i$-th superpixel's label, and $\beta$ is a parameter.

By minimizing the two energy functions above, we have,
\begin{equation}
\setlength{\abovedisplayskip}{2pt}
\setlength{\belowdisplayskip}{2pt}
\label{cross1}
\boldsymbol{l}^d_{(t+1)} = \arg\min_{\boldsymbol{l}} E^d_{(t+1)}(\boldsymbol{l}) = (L^d_{(t+1)} + \beta I)^{-1} \boldsymbol{l}^c_{(t)}, 
\end{equation}
\begin{equation}
\setlength{\abovedisplayskip}{2pt}
\setlength{\belowdisplayskip}{2pt}
\label{cross2}
\boldsymbol{l}^c_{(t+1)} = \arg\min_{\boldsymbol{l}} E^c_{(t+1)}(\boldsymbol{l}) = (L^c_{(t+1)} + \beta I)^{-1} \boldsymbol{l}^d_{(t)}, 
\end{equation}
where $L$ is the Laplacian matrix. $\boldsymbol{l}^c_{(0)}$ and $\boldsymbol{l}^d_{(0)}$ are set to the results of the Saliency Game stated in Section~\ref{section:my}. After $T$ rounds, the iteration converges and the final saliency map is obtained as:
\begin{equation}
\setlength{\abovedisplayskip}{2pt}
\setlength{\belowdisplayskip}{2pt}
\label{final}
S = \rho_1 \cdot \boldsymbol{l}^c_{(T)} + \rho_2 \cdot \boldsymbol{l}^d_{(T)}, 
\end{equation}
where $\rho_1$ and $\rho_2$ control the weight of the two results.

As shown in Figure~\ref{fig:5}(f), through the Iterative Random Walk, information from the color space helps cut the whole salient object clearly. Semantic information from deep features helps locate the target object accurately. Also, objects that could not be detected in one feature space can be detected with the help of results from the other feature space. 

\section{Experiments and Results}
In this section, we evaluate the proposed method on 6 benchmark datasets: ECSSD~\cite{Yan2013Hierarchical} (1000 images), PASCAL-S~\cite{Li2014The} (850 images), MSRA-5000~\cite{Liu2011Learning} (5000 images), HKU-IS~\cite{Li2015Visual} (4447 images), DUT-OMRON~\cite{yang2013saliency} and SOD~\cite{Movahedi2010Design}. 

We compare our algorithm with 11 state-of-the-art methods including BL~\cite{tong2015bootstrap}, BSCA~\cite{Qin_2015_CVPR}, DRFI~\cite{jiang2013salient}, DSR~\cite{li2013saliency}, HS~\cite{Yan2013Hierarchical}, LEGS~\cite{Wang2015Deep}, MCDL~\cite{zhao2015saliency}, MR~\cite{yang2013saliency}, RC~\cite{cheng2015global}, wCO~\cite{zhu2014saliency}, and KSR~\cite{Tiantian2016Kernelized}. Results of different methods are provided by authors or achieved by running available codes.

\subsection{Parameter setting}
\label{parameter}
All parameters are set once fixed over all the datasets. We segment an image into 100, 150, 200, and 250 superpixels (i.e., 4 segmentation image), run the algorithm on each map, and average the four outputs to form the final saliency map. $\sigma$ is set to 0.1 and $\epsilon$ is set to $10^{-4}$. The parameters controlling the weight of each term in the payoff function (Eqn.~\ref{payoff}) are set to $\lambda_1 = 2.1 \times 10^{-6}, \lambda_2 = 9 \times 10^{-7}$, respectively. $\alpha$ in Eqn.~\ref{supp} is set to 0.007. $\beta$ in Eqn.~\ref{cross1} and Eqn.~\ref{cross2} is set to 1. In Eqn.~\ref{final}, we set $T=20$, $\rho_1=0.3$ and $\rho_2=0.7$. 

The proposed method is implemented in MATLAB on a PC with a 3.6GHz CPU and 32GB RAM. It takes about 2.3 seconds to generate a saliency map, excluding the time for deep feature extraction and superpixel segmentation.

\subsection{Evaluation metrics}
We use precision-recall curve, F-measure curve, F-measure and AUC to quantitatively evaluate the experimental results. The precision value is defined as the ratio of salient pixels correctly assigned to all salient pixels in the map to be evaluated, while the recall value corresponds to the percentage of the detected salient pixels with respect to all salient pixels in the ground-truth map. The F-measure is an overall performance indicator computed by the weighted harmonic of precision and recall. We set $\beta^2 = 0.3$ as suggested in~\cite{Achanta2009Frequency} to emphasize the precision. 

Given a saliency map with intensity values normalized to the range of 0 and 1, a series of binary maps are produced by using several fixed thresholds in $[0, 1]$. We compute the precision/recall pairs of all the binary maps to plot the precision-recall curves and the F-measure curves. As suggested in~\cite{Achanta2009Frequency}, we use twice the mean value of the saliency maps as the threshold to generate binary maps for computing F-measure. Notice that some works have reported slightly different F-measures using different thresholds.
\begin{figure}[t]
\begin{center}
\includegraphics[width=1\linewidth]{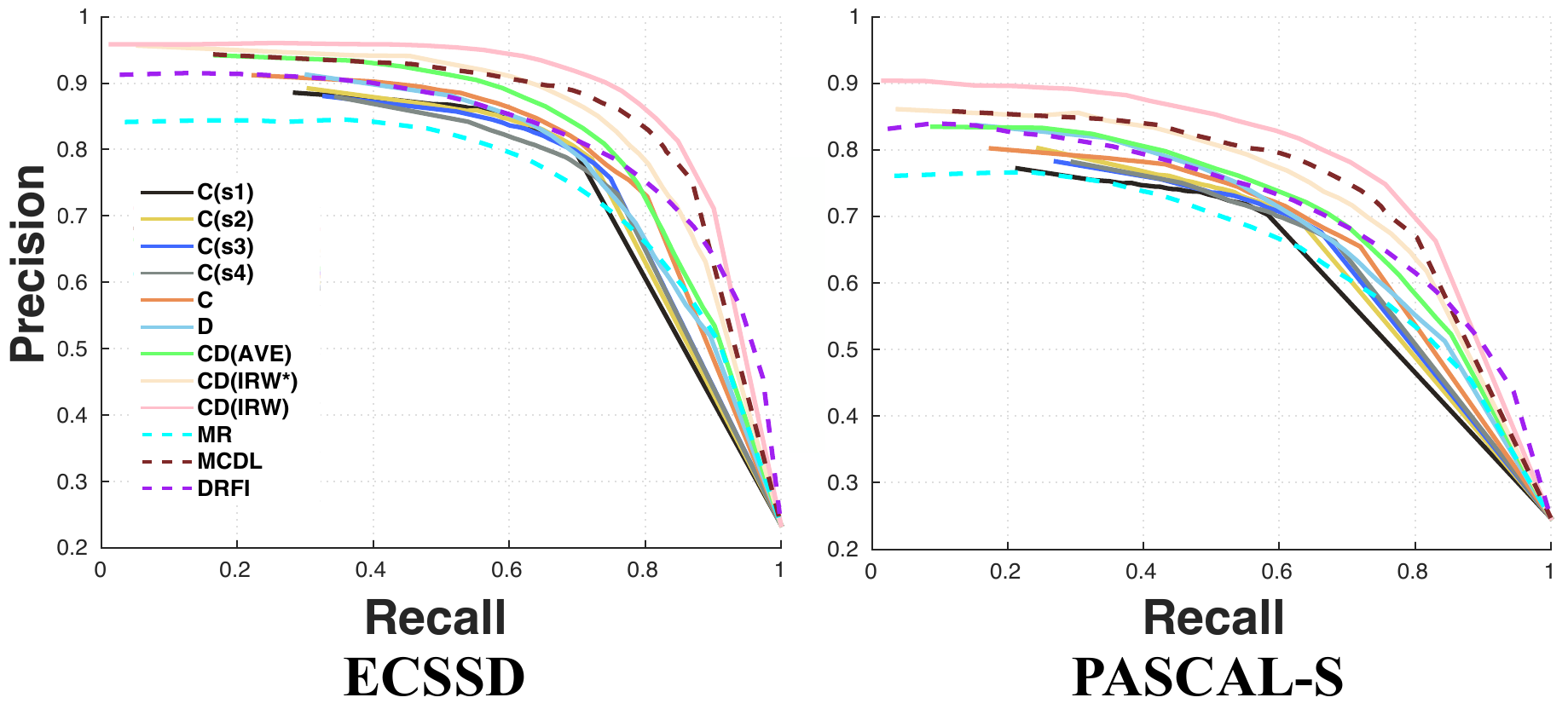}
\end{center}
\vspace{-10pt}
\caption{Effect of each step of the algorithm. Left) PR curves on ECSSD dataset. Right) PR curves on PASCAL-S dataset. The performances are compared with three existing methods MCDL~\cite{zhao2015saliency}, DRFI~\cite{jiang2013salient} and MR~\cite{yang2013saliency}. C(s1), C(s2), C(s3), C(s4): Color feature Saliency Game with four different scales of superpixels. C: Color feature Saliency Game over scales. D: Deep feature Saliency Game averaged over scales. CD(AVE): a weighted sum of C and D. CD(IRW): Saliency Game refined by the Iterative Random Walk with metric cross fusion. SG(IRW*): Saliency Game refined by the Iterative Random Walk without metric cross fusion.}
\label{fig:6}
\end{figure}

\subsection{Algorithm validation}
To demonstrate the effectiveness of each step of our algorithm, we test the proposed Saliency Game and the Iterative Random Walk (with and without metric fusion) separately on ECSSD and PASCAL-S datasets. PR curves in Figure~\ref{fig:6} show that:
\begin{itemize}
\item The proposed saliency game algorithm achieves favorable performance. Note that even when using only simple color features, as a fully unsupervised method, our proposed Saliency Game (C in Figure~\ref{fig:6}) algorithm is comparable with supervised methods.
\item The Iterative Random Walk improves performance in both deep feature space (D in Figure~\ref{fig:6}) and color space CD(IRW). Comparing results of Iterative Random Walk CD(IRW), Iterative Random Walk without metric fusion CD(IRW*), and weighted sum of results in the two feature spaces CD(ave), demonstrates the advantage of cross penalization and metric fusion in the Iterative Random Walk.
\end{itemize}

\begin{figure}[t]
\begin{center}
  \includegraphics[width=1\linewidth]{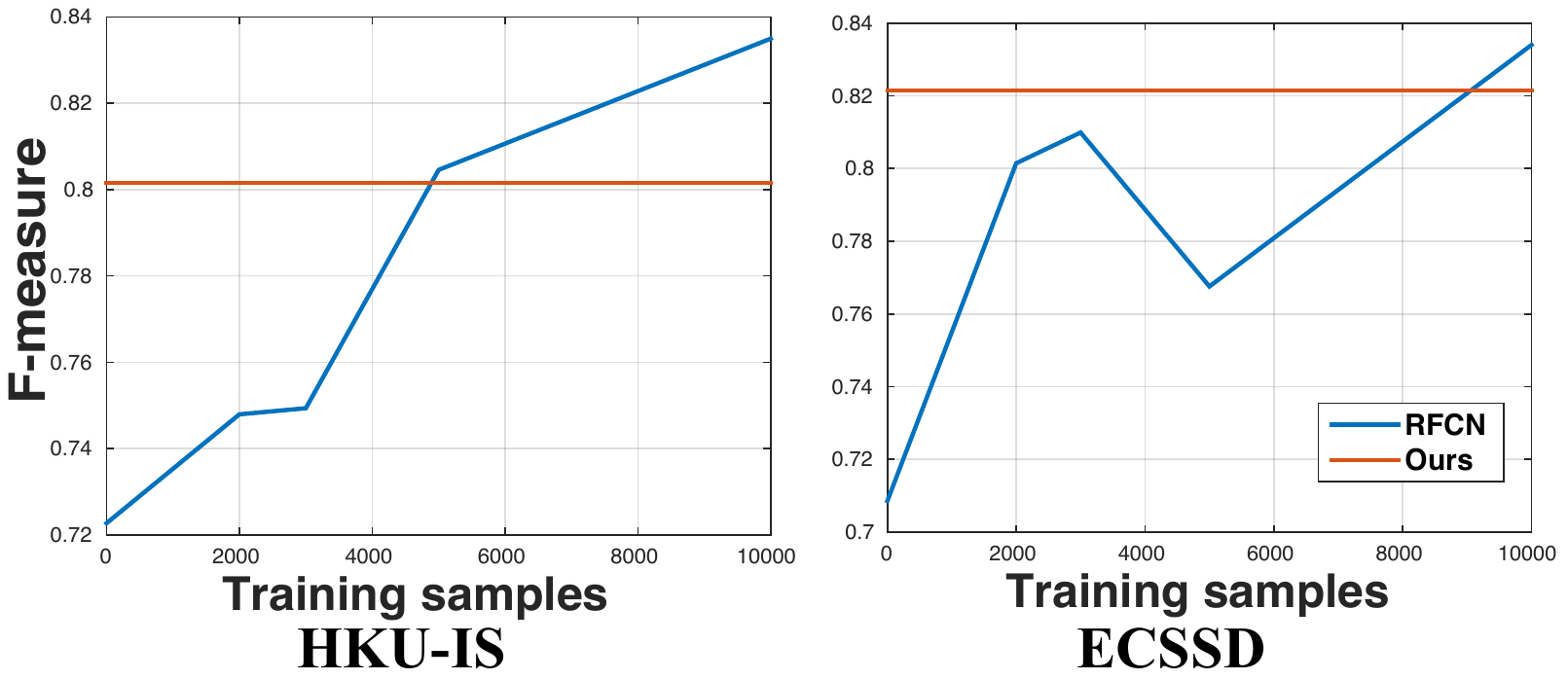}
\end{center}
\vspace{-10pt}
  \caption{Comparison in terms of F-measure between our method and RFCN fine-tuned on different number of training samples, evaluated on ECSSD and HKU-IS datasets.}
\label{fig:7}
\end{figure}

In addition, as an unsupervised approach, our method is economical and practical. Although some deep learning based methods outperform ours in few cases, a lot of time and a large number of training samples are required to assure their effectiveness. Otherwise, performance of these methods might not be as well as ours. To demonstrate this, we show comparison in terms of F-measure between our method and RFCN~\cite{Wang2016Saliency} fine-tuned on different number of training samples in Figure~\ref{fig:7}. RFCN is a recently proposed deep learning based method that achieved excellent performance. However, it can be seen from the figure that RFCN does not do well without fine-tuning. Its F-measure increases as the number of training samples grows. Our method is equivalent to RFCN fine-tuned on about 5000-9000 images.

\begin{table}[h]
\begin{center}
\mbox{\scriptsize
\begin{tabular}{|c|c|c|c|c|c|c|}
\hline
    dataset                 &BL  	 &DSR  	  &HS 	   &MR                     &RC      &wCO      \\
\hline
    ECSSD                   &0.6838  &0.6618  &0.6347  &0.6905                 &0.4560  &0.6764\\
    PASCAL-S                &0.5742  &0.5575  &0.5314  &0.5863                 &0.4039  &0.5999\\
    MSRA                    &0.7840  &0.7841  &0.7671  &\color{green}{0.8041}  &0.5754  &0.7937\\
    HKU-IS                  &0.6597  &0.6774  &0.6359  &0.6550                 &0.5008  &0.6770\\
    SOD                   &0.5723  &0.5962  &0.5212  &0.5695                 &0.4184  &0.5987\\
    DUT-OMRON 				&0.4989  &0.5243  &0.5108  &0.5280                 &0.4058  &0.5277\\
\hline
                       &BSCA    &\textbf{DRFI}&\textbf{MCDL}&\textbf{LEGS}&\textbf{KSR}&Ours\\
\hline
    ECSSD              &0.7046  &0.7329  &\color{green}{0.7959}  &0.7851  &0.7817               &\color{red}0.8215\\
    PASCAL-S           &0.6006  &0.6182  &0.6912                 &-       &\color{green}{0.7039}  &\color{red}0.7062\\
    MSRA               &0.7934  &-       &-                      &-       &-                    &\color{red}0.8666\\
    HKU-IS             &0.6545  &0.7219  &\color{green}{0.7573}  &0.7229   &0.7468              &\color{red}0.8015\\
    SOD              &0.5835  &0.6470  &0.6772  &\color{green}{0.6834}  &0.6679               &\color{red}{0.6896}\\
    DUT-OMRON 			&0.5091  &0.5505  &\color{red}{0.6250 } &0.5916      &0.5911  &\color{green}{0.5981}\\
\hline
\end{tabular}
}\end{center}
\caption{F-measure scores. The best and the second best results are shown in red and green, respectively. Supervised methods are marked in bold.}
\label{tab1}
\end{table}

\begin{table}[h]
\begin{center}
\mbox{\scriptsize
\begin{tabular}{|c|c|c|c|c|c|c|}
\hline
    dataset                 &BL  					&DSR  	 &HS  	  &MR	   &RC  	&wCO      \\
\hline
    ECSSD                   &0.9143  				&0.8619  &0.8838  &0.8827  &0.8342  &0.8814\\
    PASCAL-S                &0.8671  				&0.8118  &0.8362  &0.8259  &0.8139  &0.8482\\
    MSRA                    &\color{green}{0.9535}  &0.9382  &0.9279  &0.9267  &0.8951  &0.9360\\
    HKU-IS                  &0.9140  				 &0.9008  &0.8782  &0.8611  &0.8530  &0.8952\\
    SOD	                    &\color{green}{0.8503}  				 &0.8208  &0.8145  &0.7903  &0.7924  &0.8026\\
    DUT-OMRON               &0.8778  				 &0.8787  &0.8586  &0.8447  &0.8476  &0.8846\\
\hline
                       &BSCA    &\textbf{DRFI}	&\textbf{MCDL}&\textbf{LEGS}&\textbf{KSR}&Ours\\
\hline
    ECSSD              &0.9176  &\color{red}{0.9404}    &0.9186  &0.9235  &0.9268      &\color{green}0.9272\\
    PASCAL-S           &0.8665  &\color{green}{0.8950}  &0.8699  &-       &\color{red}0.9012      &0.8724\\
    MSRA               &0.9484  &-       				&-       &-       &-           &\color{red}0.9583\\
    HKU-IS             &0.9140  &\color{red}{0.9435}  	&0.9175  &0.9026  &0.9099      &\color{green}0.9183\\
    SOD             &0.8358  &\color{red}{0.8813}  &0.8163  &0.8268  &0.8403  &0.8481\\
	DUT-OMRON          &0.8779  &\color{red}{0.9157}  	&\color{green}{0.9014}  		&0.8841  &0.8921  &0.8869\\
\hline
\end{tabular}
}\end{center}
\caption{AUC scores. Supervised methods are marked in bold.}
\label{tab2}
\end{table}
\subsection{Comparison with state-of-the-Art methods}
\begin{figure*}[htbp]
\begin{center}
\includegraphics[width=1\linewidth]{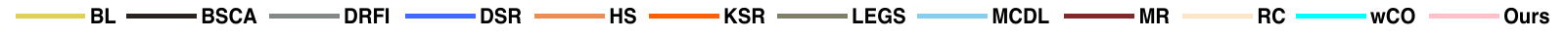}\\
\includegraphics[width=1\linewidth]{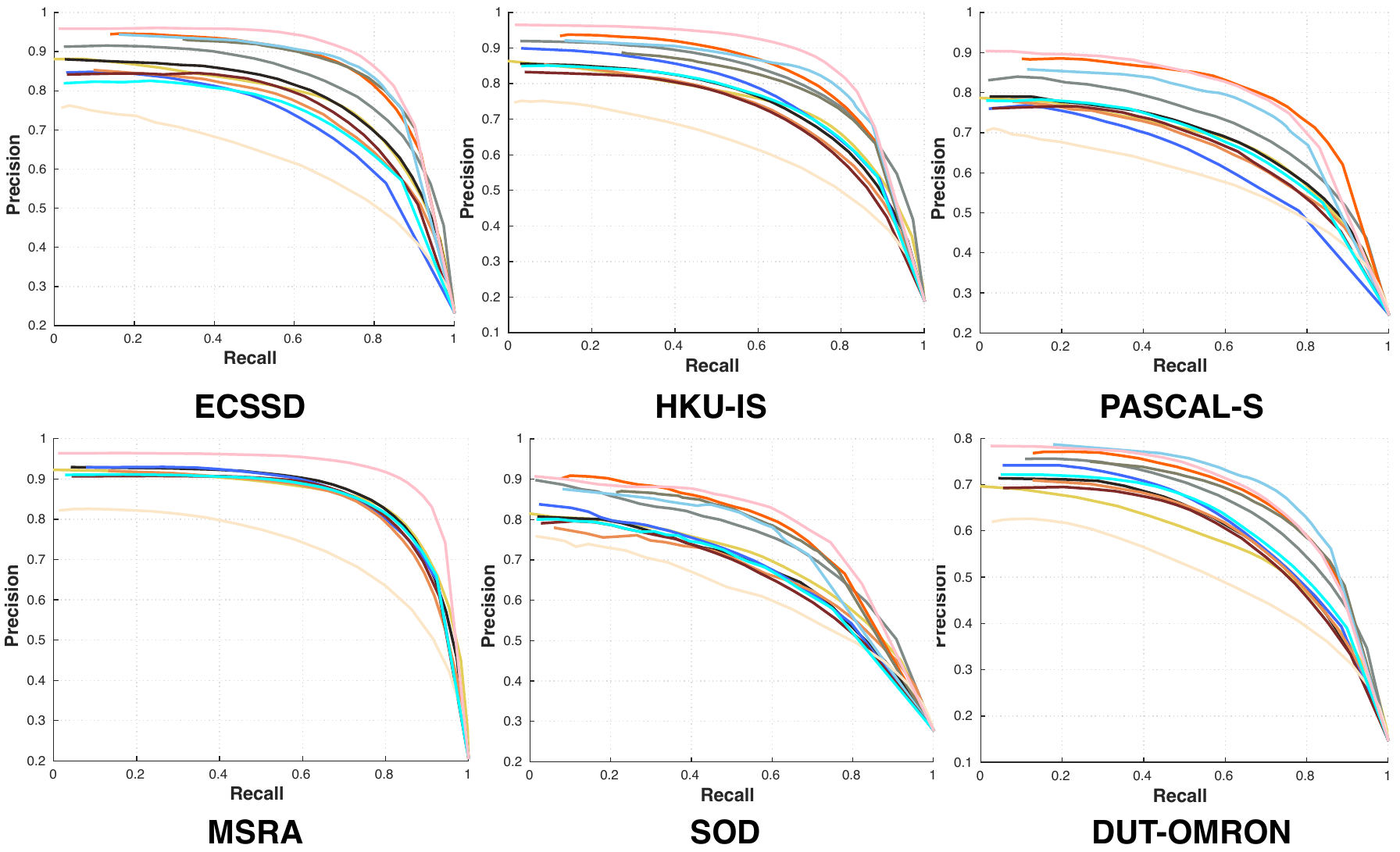}
\end{center}
\vspace{-10pt}
\caption{Quantitative comparison of models in terms of PR curves.}
\label{fig:8_1}
\end{figure*}

\begin{figure*}[htbp]
\begin{center}
\includegraphics[width=1\linewidth]{label.pdf}\\
\includegraphics[width=1\linewidth]{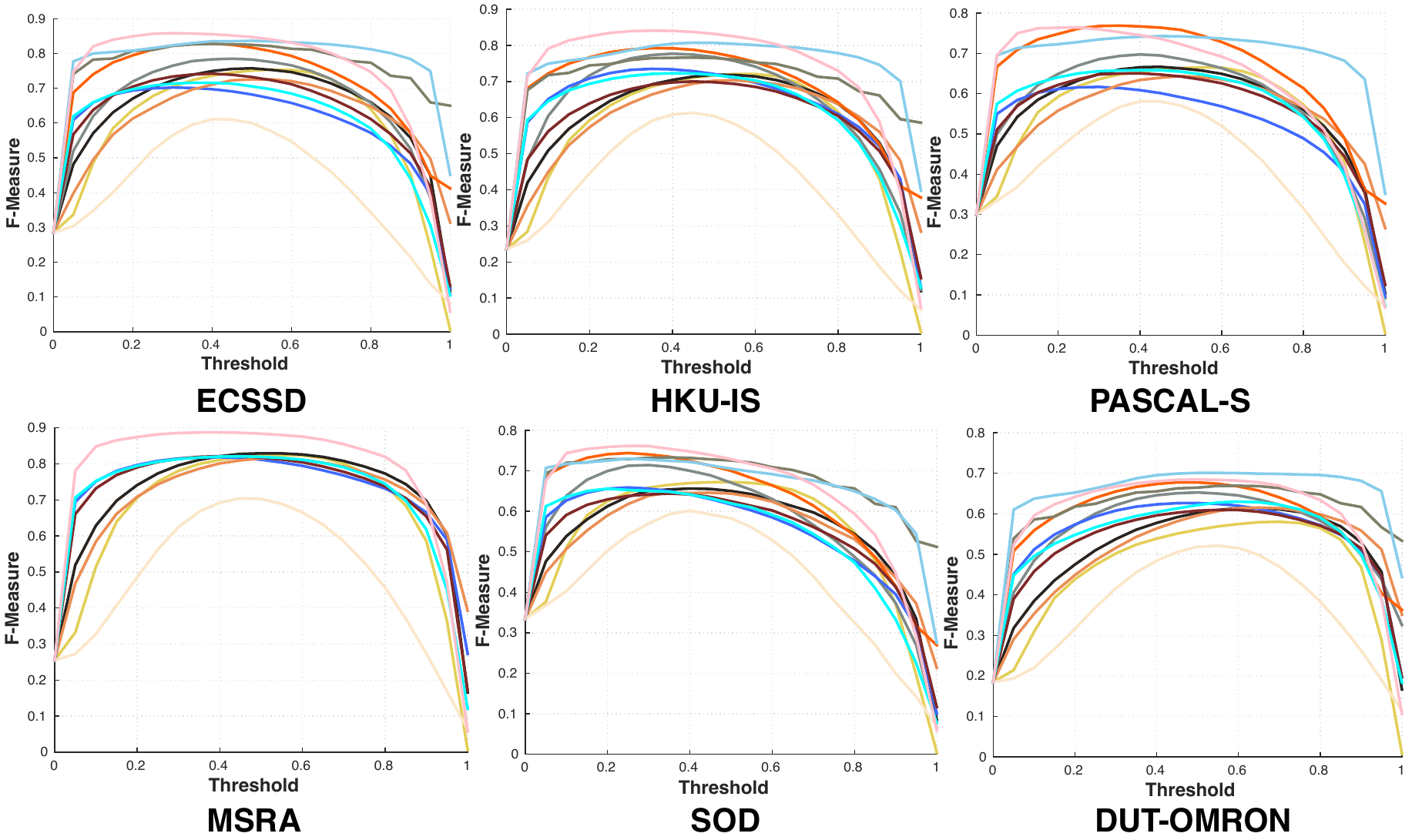}
\end{center}
\vspace{-10pt}
\caption{Quantitative comparison of models in terms of F-measure.}
\label{fig:8_2}
\end{figure*}

\begin{figure*}[htbp]
\begin{center}
\includegraphics[width=\linewidth]{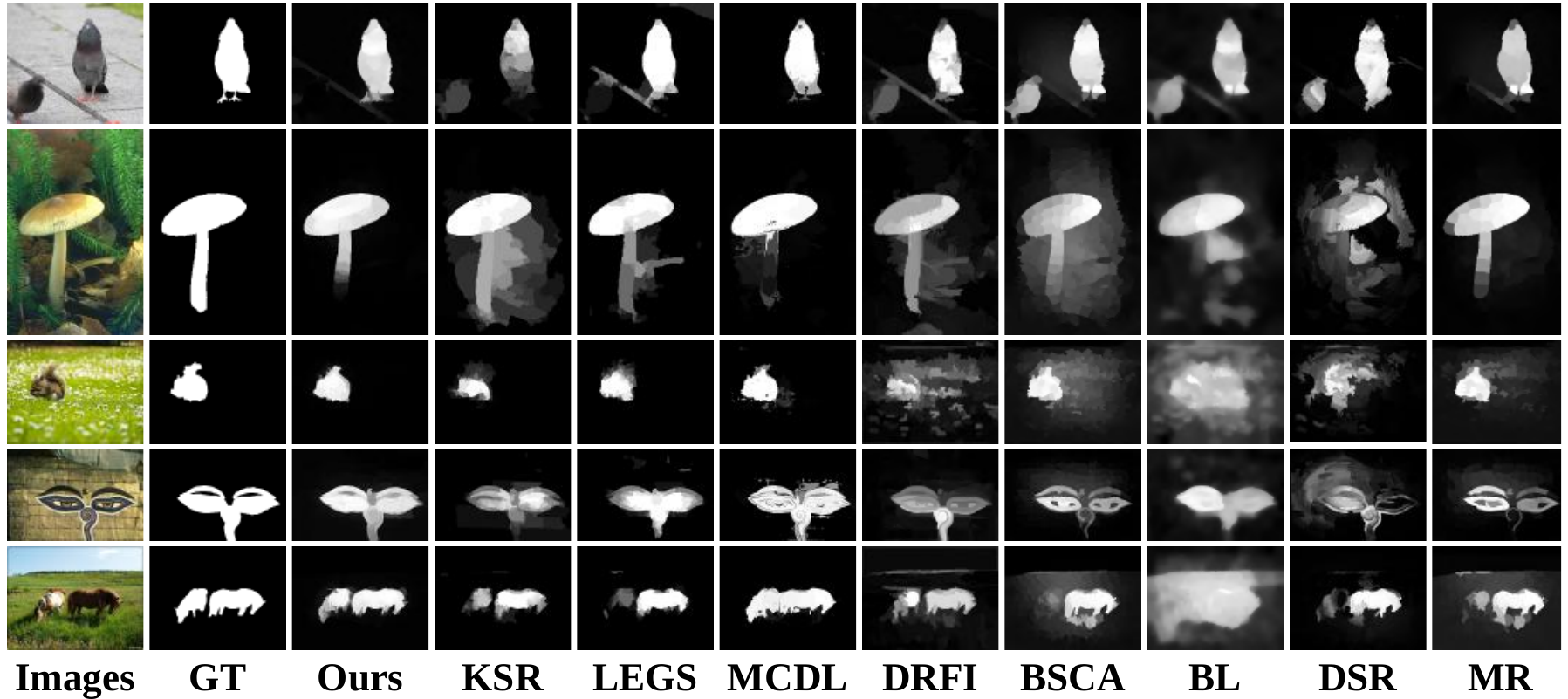}
\end{center}
\vspace{-10pt}
\caption{Qualitative comparison of our methods and state-of-the-art methods on ECSSD dataset.}
\label{cmp_ecssd}
\end{figure*}

\begin{figure*}[htbp]
\begin{center}
\includegraphics[width=\linewidth]{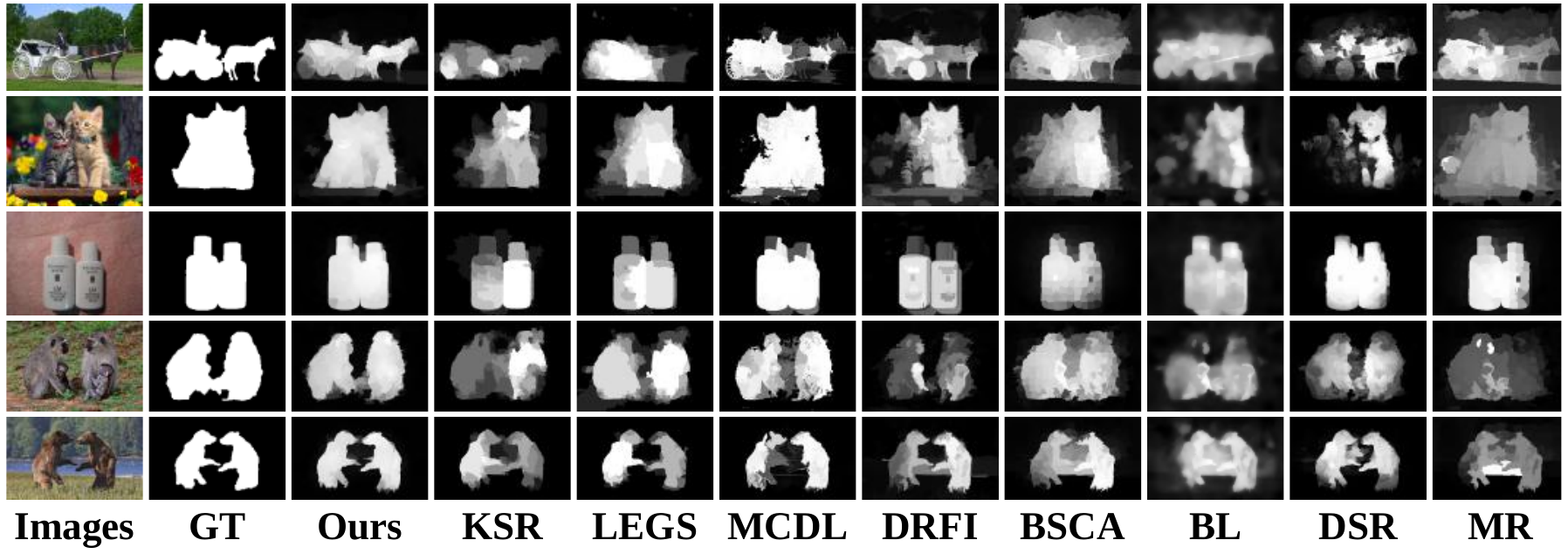}
\end{center}
\vspace{-10pt}
\caption{Qualitative comparison of our methods and state-of-the-art methods on HKU-IS dataset.}
\label{cmp_hku}
\end{figure*}

\begin{figure*}[t]
\begin{center}
\includegraphics[width=\linewidth]{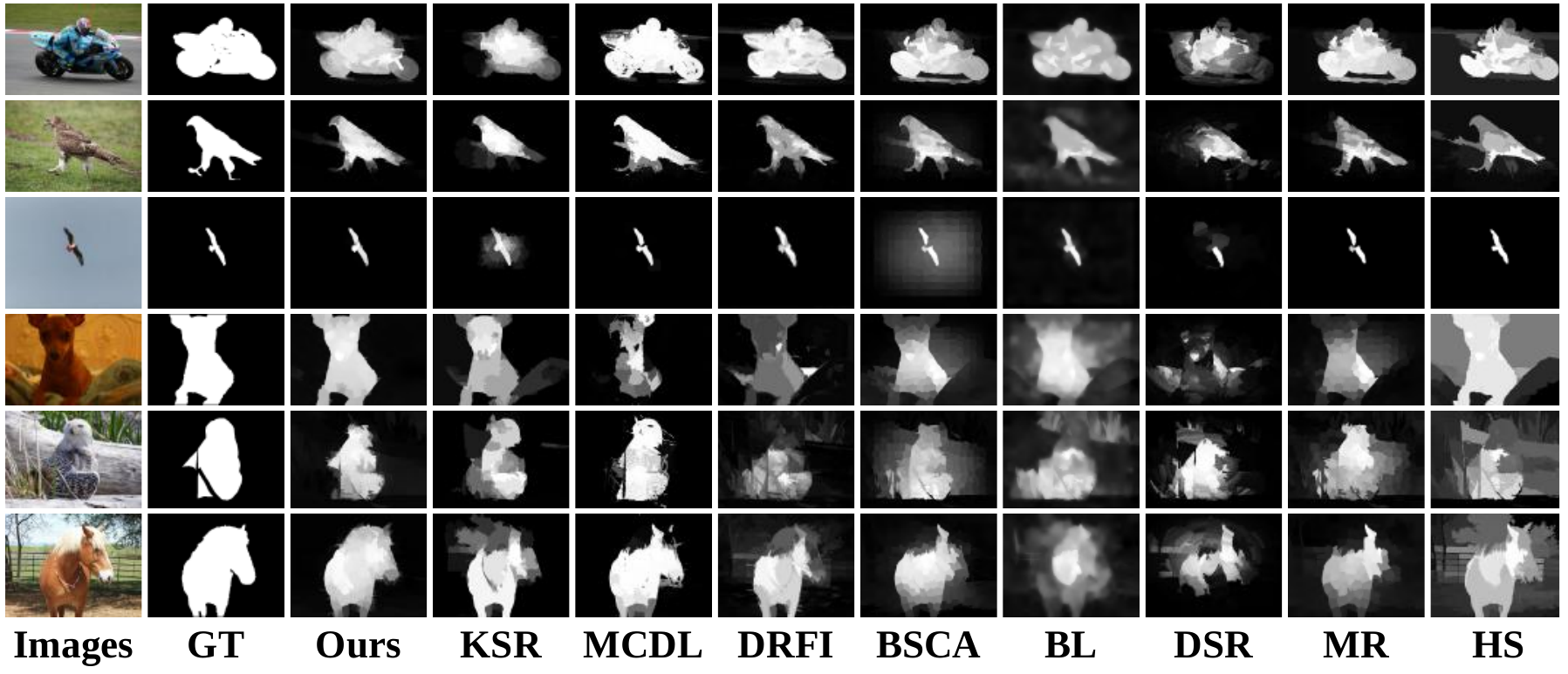}
\end{center}
\vspace{-10pt}
\caption{Qualitative comparison of our methods and state-of-the-art methods on PASCAL-S dataset.}
\label{cmp_pascals}
\end{figure*}

\begin{figure*}[t]
\begin{center}
\includegraphics[width=\linewidth]{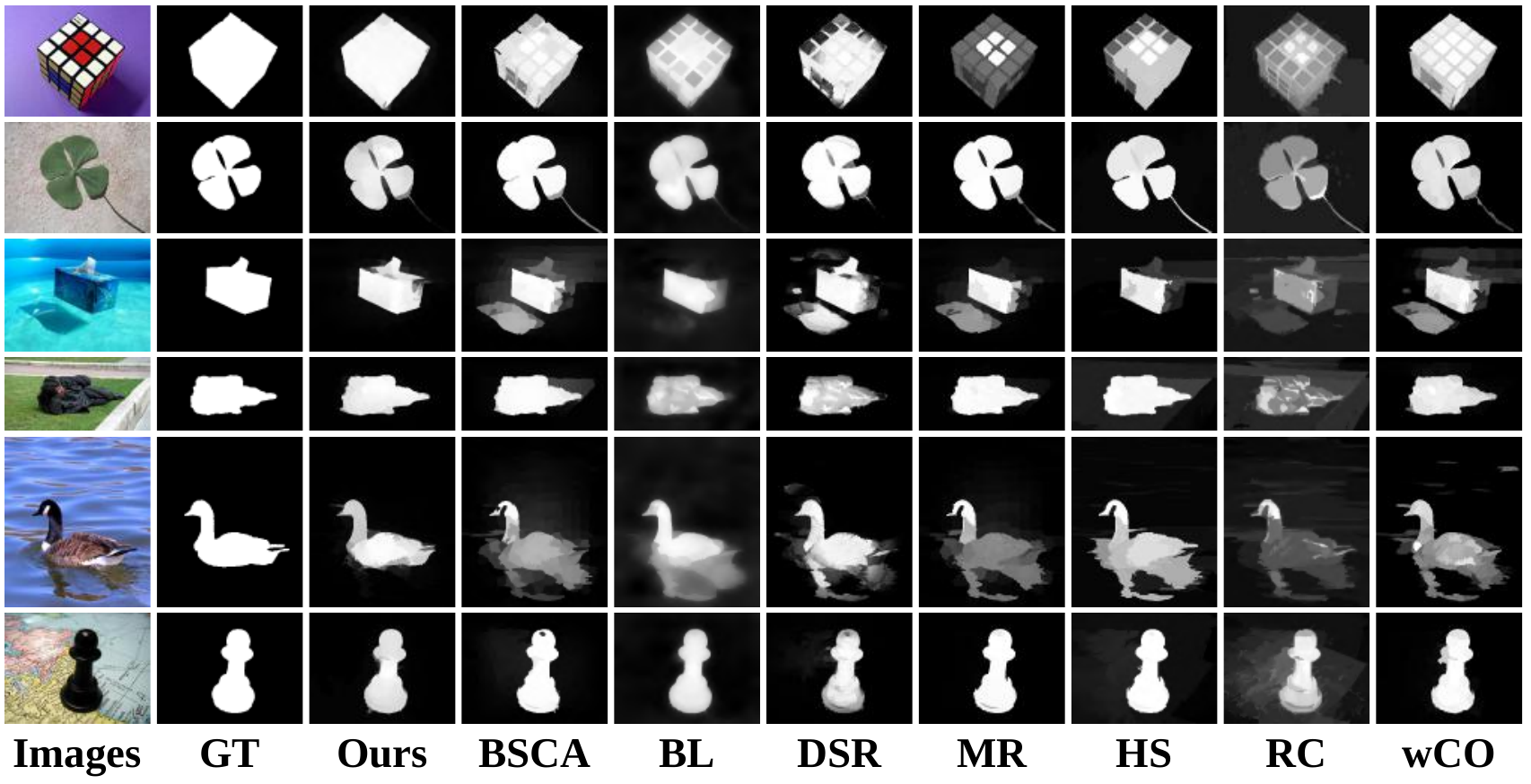}
\end{center}
\vspace{-10pt}
\caption{Qualitative comparison of our methods and state-of-the-art methods on MSRA dataset.}
\label{cmp_msra}
\end{figure*}

As is shown in Figure~\ref{fig:8_1}, Figure~\ref{fig:8_2}, Table~\ref{tab1} and Table~\ref{tab2}, our proposed method compares favorably against 11 state-of-the-art approaches over six different datasets. Among models, BL, BSCA, DSR, HS, MR, RC, wCO are unsupervised methods. DRFI, LEGS, MCDL, KSR are supervised methods. DRFI learns a random forest regressor, LEGS and MCDL train a convolutional neural network, KSR learns a classifier and a subspace projection to rank object proposals based on R-CNN features. For a fair comparison, we do not provide evaluation results of DRFI, LEGS, MCDL, and KSR methods on MSRA-5000 dataset since these methods all randomly select images from this dataset for training. Further, since LEGS also selects images from PASCAL-S dataset, we do not show its performance over the PASCAL-S dataset. Visual comparison of the proposed method against state-of-the-art on different datasets is shown in Figure~\ref{cmp_ecssd}, \ref{cmp_hku}, \ref{cmp_pascals}, \ref{cmp_msra}.

\subsection{Sensitivity analysis}
\begin{figure*}[htbp]
\begin{center}
\includegraphics[width=1\linewidth]{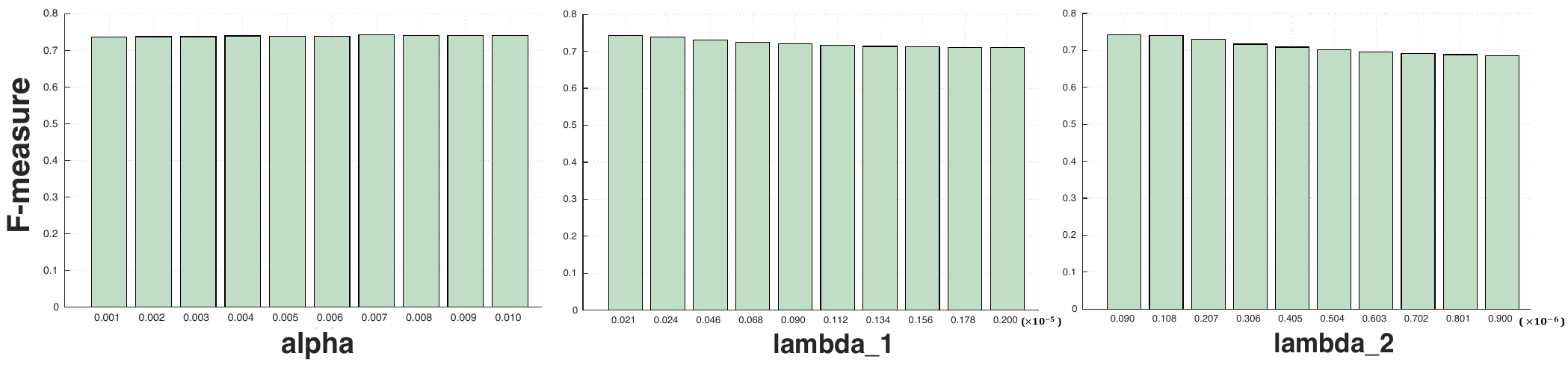}
\end{center}
\vspace{-10pt}
\caption{Sensitivity of the Saliency Game to parameters $\alpha$, $\lambda_1$and $\lambda_2$, evaluated on ECSSD dataset.}
\label{sens1}
\end{figure*}

\begin{figure}[htbp]
\begin{center}
\includegraphics[width=1\linewidth]{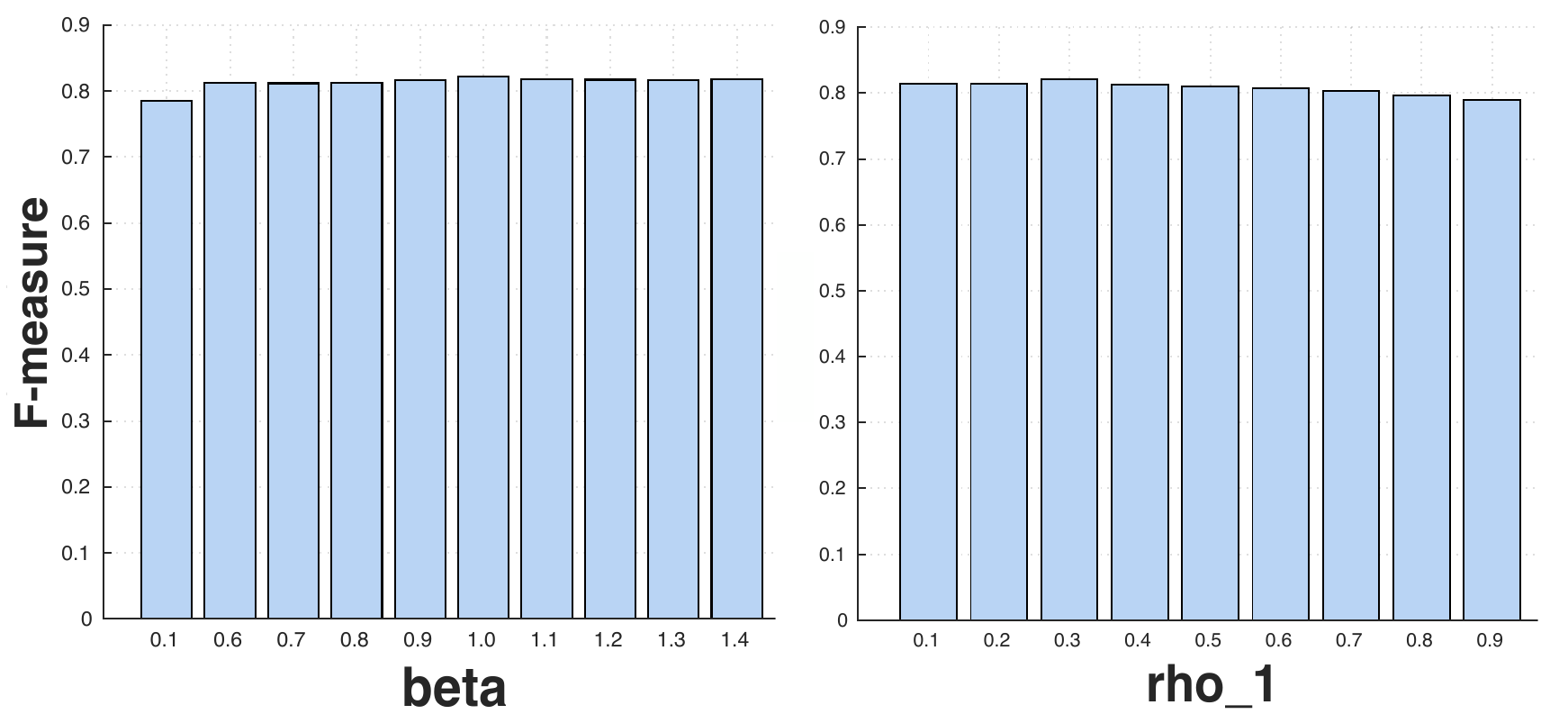}
\end{center}
\vspace{-10pt}
\caption{Sensitivity of the Iterative Random Walk to parameters $\beta$ and $\rho_2$, evaluated on ECSSD dataset.}
\label{sens2}
\end{figure}

In this section, we test sensitivity of the proposed Saliency Game to parameters $\lambda_1$, $\lambda_2$, $\alpha$ and sensitivity of the Iterative Random Walk to parameters $\beta$ and $\rho_1$. As shown in Figure~\ref{sens1} and \ref{sens2}, the performance in terms of F-meansure score almost keeps the same when varying the parameters a little, so the proposed method is not sensitive to these parameters.

\subsection{Equilibria}
\begin{figure}[htbp]
\label{othereq}
\begin{center}
\includegraphics[width=1\linewidth]{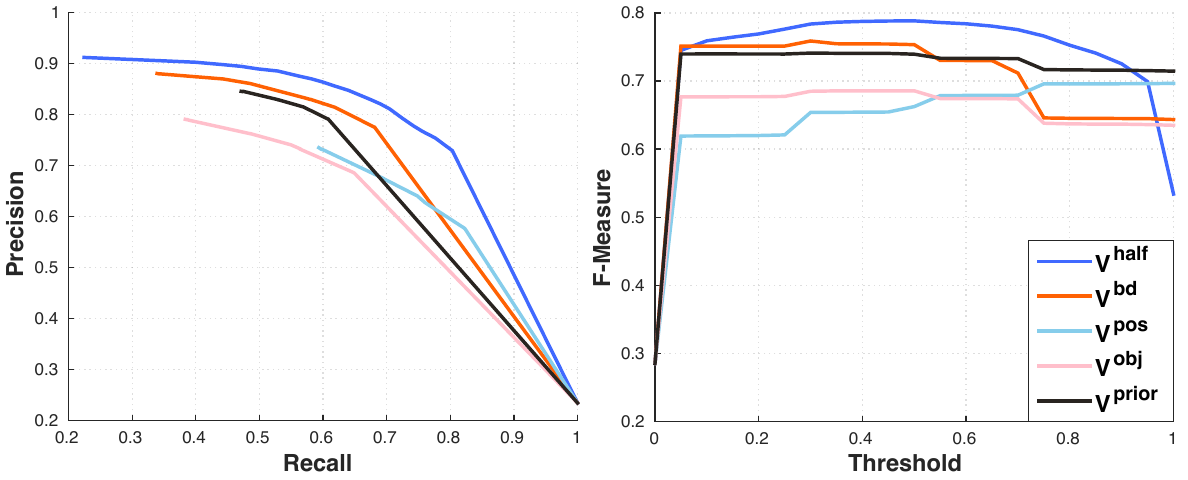}
\end{center}
\vspace{-10pt}
\caption{Comparison of different saliency detection results corresponding to five different Nash equilibria (evaluated on ECSSD dataset). $V^{half}$, $V^{pos}$, $V^{obj}$ , $V^{bd}$ and $V^{prior}$: Saliency detection results corresponding to five Nash equilibria reached by Replicator Dynamics starting from initial state $V^{half}$, $V^{pos}$, $V^{obj}$ , $V^{bd}$ and $V^{prior}$, respectively. }
\end{figure}

The proposed Saliency Game is a special category of games named polymatrix games~\cite{Howson1972Equilibria}, where each player plays a two-player game against each other and his payoff is then the sum of the payoffs from each of the two-player games~\cite{Deligkas2016An}. Howson \etal~\cite{Howson1972Equilibria} showed that every polymatrix game has at least one equilibrium. Therefore, the proposed Saliency Game also has at least one, but could have more than one equilibria. Replicator Dynamics is invoked to find a Nash equilibrium of the game, in which different Nash equilibria might be reached if the initial state $\boldsymbol{z}_i(0)$ is set to different interior points of $\Delta$. Empirically, we find that $\forall i \in \mathcal{I}, \boldsymbol{z}_i(0)=(0.5, 0.5)$ is a good initialization leading to plausible saliency detection. In this section, we show the saliency detection results corresponding to other four Nash equilibria, reached by Replicator Dynamics starting from four different interior points of $\Delta$. We denote the initial state used in the paper as $V^{half}$, and the other four initial states as $V^{bd}$, $V^{pos}$, $V^{obj}$, and $V^{prior}$. Each of them is a $2\times N$ matrix, where the $i$-th column vector (denoted as $\boldsymbol{v}_i^{bd}$, $\boldsymbol{v}_i^{pos}$, $\boldsymbol{v}_i^{obj}$, $\boldsymbol{v}_i^{prior}$, respectively) corresponds to the mixed strategy of superpixel $i$. Variables $\boldsymbol{v}_i^{bd}$, $\boldsymbol{v}_i^{pos}$, $\boldsymbol{v}_i^{obj}$ and $\boldsymbol{v}_i^{prior}$ are set as follows:
\begin{equation}
\boldsymbol{v}_i^{bd,1} = 
\begin{cases}
0.4 &\mbox{if}~i \in \mathcal{B}\\
0.5 &\mbox{otherwise}
\end{cases}, 
\boldsymbol{v}_i^{bd,0} = 1-\boldsymbol{v}_i^{bd,1};
\end{equation}
\begin{equation}
\boldsymbol{v}_i^{pos,1} = N \cdot pos_i(1), 
\boldsymbol{v}_i^{pos,0} = N \cdot pos_i(0);
\end{equation}
\begin{equation}
\boldsymbol{v}_i^{obj,1} = N \cdot obj_i(1), 
\boldsymbol{v}_i^{obj,0} = N \cdot obj_i(0);
\end{equation}
\begin{equation}
\boldsymbol{v}_i^{prior,1} = prior_i, 
\boldsymbol{v}_i^{prior,0} = 1-prior_i;
\end{equation}
where $prior_i$ is the saliency value of superpixel $i$ computed by another saliency detection method. In this experiment, we use MR~\cite{yang2013saliency} model to compute $prior_i, \forall i \in \mathcal{I}$. We try four different initial states to test whether inducing prior knowledge into the initial state leads to a better saliency detection result. Each of the five different Nash equilibria corresponds to a saliency detection result. We show the quantitative comparison of the five different results in terms of F-measure curves and PR curves in Figure~\ref{othereq}. It can be seen that the initial state  $V^{half}$ without any prior knowledge, which is adopted in the paper, leads to the best saliency detection.  


\section{Summary and Conclusion}
We propose a novel saliency detection algorithm. Firstly, we formulate a Saliency Game among superpixels, and a saliency map is generated according to each region's strategy in the Nash equilibrium of the proposed Saliency Game. Secondly, an iterative random walk that combines a deep feature and a color feature is constructed to refine the saliency maps generated in the last step. Extensive experiments over four benchmark datasets demonstrate that the proposed algorithm achieves favorable performance against state-of-the-art methods. The sensitivity analysis shows the robustness of the proposed method to parameter changes.

Different from most previous methods that formulate saliency detection as minimizing one single energy function, the game-theoretic approach can be regarded as maximizing many competing objective functions simultaneously. The game equilibrium automatically provides a trade-off. This seems very natural for attention modeling and saliency detection, as also features and objects compete to capture our attention. In addition, compared with CNN based saliency detection methods which need to be trained on images with pixel-level masks as ground truth, the proposed method extracts features from a pre-trained CNN and combines them with color features in an unsupervised manner. This provides an efficient complement to CNNs that does on par with models that need labeled training data. Hopefully, our approach will encourage future models that can utilize both labeled and unlabeled data. 

%



\section*{Acknowledgment}

The authors would like to thank...

\ifCLASSOPTIONcaptionsoff
  \newpage
\fi



%
\bibliographystyle{ieee}
\bibliography{reference.bib}

%

\begin{IEEEbiography}[{\includegraphics[width=1in,height=1.25in,clip,keepaspectratio]{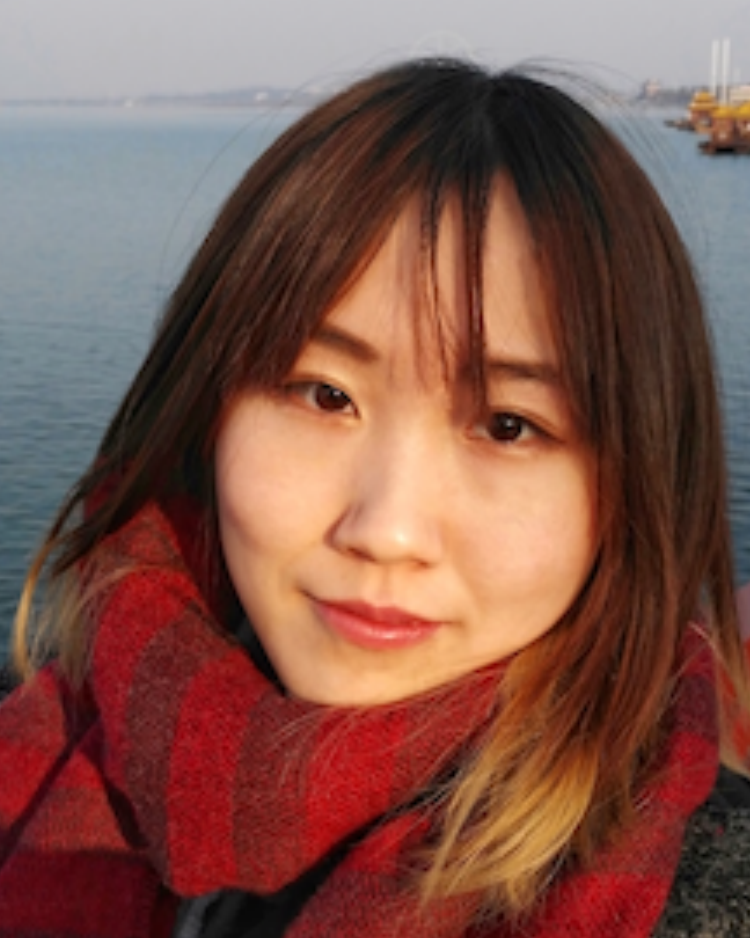}}]{Yu Zeng}
received the B.S. degree in Electronic and Information Engineering from Dalian University of Technology (DUT), Dalian, China, in 2017. Her current research interests include image processing and computer vision with focus on image statistics and saliency detection.
\end{IEEEbiography}
\begin{IEEEbiography}[{\includegraphics[width=1in,height=1.25in,clip,keepaspectratio]{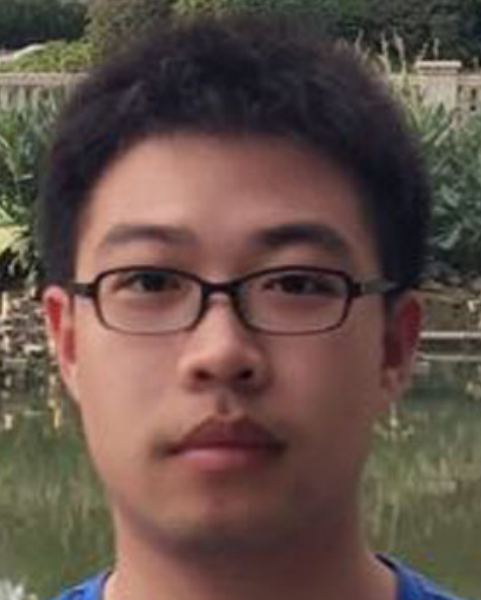}}]{Mengyang Feng}
received B.E. degree in Electrical and Information Engineering from Dalian University of Technology in 2015. He is currently pursuing the Ph.D. degree under the supervision of Prof. H. Lu in Dalian University of Technology. His research interest focuses on salient object detection.
\end{IEEEbiography}
\begin{IEEEbiography}[{\includegraphics[width=1in,height=1.25in,clip,keepaspectratio]{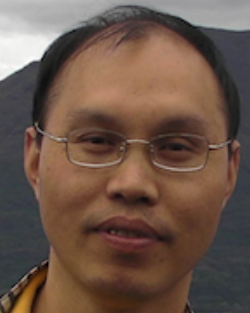}}]{Huchuan Lu}
received the Ph.D. degree in System Engineering and the M.S. degree in Signal and Information Processing from Dalian University of Technology (DUT), Dalian, China, in 2008 and 1998, respectively. He joined the faculty in 1998 and currently is a Full Professor of the School of Information and Communication Engineering, DUT. His current research interests include computer vision and pattern recognition with focus on visual tracking, saliency detection, and segmentation. He is a member of the ACM and an Associate Editor of the IEEE Transactions on Cybernetics.
\end{IEEEbiography}
\begin{IEEEbiography}[{\includegraphics[width=1in,height=1.25in,clip,keepaspectratio]{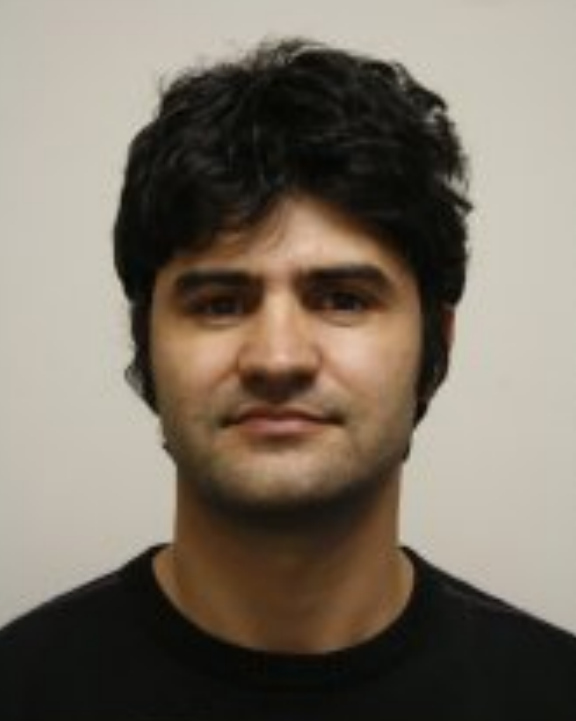}}]{Ali Borji}
received the B.S. degree in Computer Engineering from the Petroleum University of Technology, Tehran, Iran, in 2001, the M.S. degree in computer engineering from Shiraz University, Shiraz, Iran, in 2004, and the Ph.D. degree in cognitive neurosciences from the Institute for Studies in Fundamental Sciences, Tehran, Iran, in 2009. He spent four years as a Post-Doctoral Scholar with iLab, University of Southern California, from 2010 to 2014. He is currently an Assistant Professor with the University of Wisconsin, Milwaukee. His research interests include visual attention, active learning, object and scene recognition, and cognitive and computational neurosciences. 
\end{IEEEbiography}
\end{document}